\documentclass{SCIS2026}
\usepackage{listings}
\usepackage{inconsolata}
\newcommand{\eg}{e.g.}
\lstdefinestyle{customprompt}{
  basicstyle=\ttfamily\scriptsize,
  frame=single,
  breaklines=true,
  columns=flexible,
  escapeinside={(*@}{@*)} 
}

\usepackage{multirow}
\usepackage{makecell}
\usepackage{threeparttable}
\usepackage{caption}
\usepackage{graphicx}

\usepackage[table]{xcolor}
\definecolor{revisionblue}{RGB}{0, 0, 200}

\usepackage{pifont}
\newcommand{\cmark}{\ding{51}}
\newcommand{\xmark}{\ding{55}}
\newcommand{\gmark}{\textcolor{green!60!black}{\cmark}}
\newcommand{\rmark}{\textcolor{red!80!black}{\xmark}}
\begin{document}
\ArticleType{RESEARCH PAPER}
\Year{2025}
\Month{January}
\Vol{68}
\No{1}
\DOI{}
\ArtNo{}
\ReceiveDate{}
\ReviseDate{}
\AcceptDate{}
\OnlineDate{}
\AuthorMark{}
\AuthorCitation{}

\title{RS-Agent: Automating Remote Sensing Tasks through Intelligent Agent}{RS-Agent: Automating Remote Sensing Tasks through Intelligent Agent}

\author[1\dag]{Wenjia Xu}{{xuwenjia@bupt.edu.cn}}
\author[1\dag]{Zijian Yu}{}
\author[1]{Boyang Mu}{}
\author[2]{Jiuniu Wang}{}
\author[3]{Zhiwei Wei}{{trentonwei@whu.edu.cn}}
\author[1]{Mugen Peng}{}


\address[1]{State Key Laboratory of Networking and Switching Technology, Beijing University of Posts and Telecommunications, Beijing 100876, China}
\address[2]{City University of HongKong, Hong Kong 999077, China}
\address[3]{School of Geographic Sciences, Hunan Normal University, Changsha 410081, China}

\abstract{The unprecedented advancements in Multimodal Large Language Models (MLLMs) have demonstrated strong potential for language–vision interaction in remote sensing tasks such as visual question answering and scene understanding. 
However, these models are largely constrained to basic instruction-following or descriptive tasks, 
and struggle with real-world remote sensing scenarios where multi-source data, fine-grained spatial semantics, and expert knowledge are indispensable.
To address these limitations, we propose RS-Agent, a domain-adapted intelligent agent that bridges user intent and professional remote sensing workflows through structured task planning and tool orchestration. RS-Agent is built upon four key components that explicitly follow the typical workflow of remote sensing applications: a LLM-based Central Controller for user intent understanding and analytical process planning, a dynamic toolkit for tool execution, a Solution Space for task-specific expert guidance, and a Knowledge Space for domain-level knowledge support. To further enhance the performance of RS-Agent, we introduce two novel mechanisms: Task-Aware Retrieval, which improves task planning by explicitly inferring task types and retrieving expert-defined procedural solutions, rather than relying on query-level similarity or ad-hoc tool chaining; and DualRAG, a weighted dual-path retrieval-augmented generation method, which enhances the relevance and completeness of retrieved domain knowledge. RS-Agent natively supports multiple imaging modalities, including optical and SAR imagery. For SAR tasks in particular, the agent plans and orchestrates dedicated SAR processing and analysis tools into executable workflows, improving reliability and automation under underspecified requests. Extensive experiments across 9 datasets and 18 remote sensing tasks demonstrate that RS-Agent significantly outperforms state-of-the-art MLLMs, achieving over 95\% task planning accuracy and delivering superior performance in tasks such as scene classification, object counting, and remote sensing visual question answering. These results validate the effectiveness of a dedicated remote sensing agent that fuses LLM reasoning with domain expertise for geospatial intelligence. Our work presents RS-Agent as a robust and extensible framework for advancing intelligent automation in remote sensing analysis. Our code will be available at \href{https://github.com/IntelliSensing/RS-Agent}{{https://github.com/IntelliSensing/RS-Agent}}.} 


\maketitle

\section{Introduction}

\begin{figure*}[tb]
  \centering
   \includegraphics[width=0.60\linewidth]{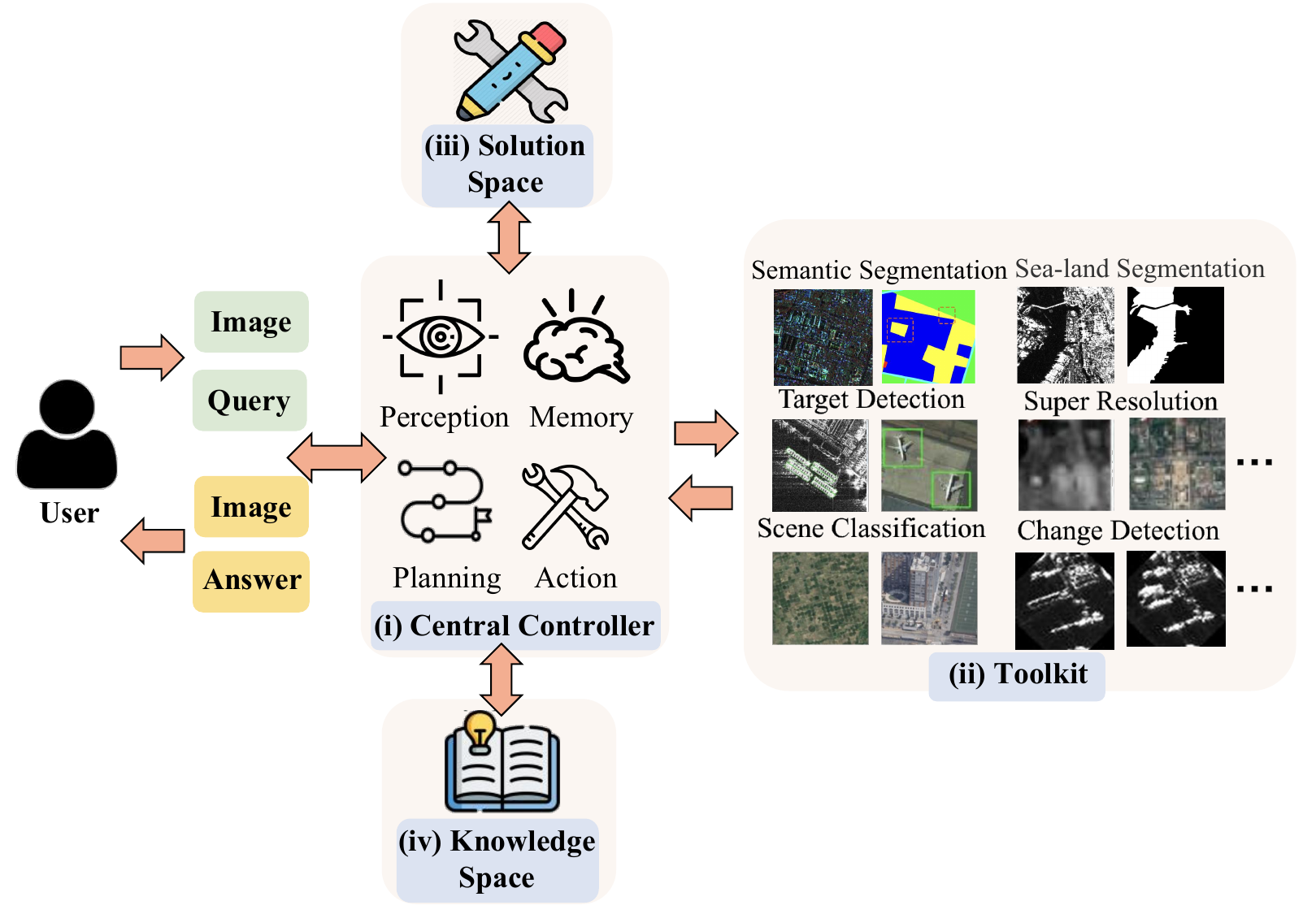}
   \caption{The schematic diagram of the RS-Agent. RS-Agent consists of four main components: a Central Controller based on LLM, which comprehends user queries, maintains historical context, plans tool usage, and aggregates results; a Solution Space that provides well-defined solutions to user-submitted problems; a Knowledge Space offering domain-specific knowledge in remote sensing; and a Toolkit integrating state-of-the-art methods for remote sensing tasks.}
   \label{fig: teaser figure}
   \vspace{-1.0em}
\end{figure*}

By aligning visual and textual information, Multi-modal Large Language Models (MLLMs)~\cite{radford2021learning,sun2023eva} have advanced remote sensing image interpretation, including captioning~\cite{tao2023general}, visual question answering~\cite{bashmal2023visual}, and scene understanding~\cite{hu2025rsgpt,kuckreja2023geochat}, making remote sensing data more accessible to non-expert users.

However, remote sensing presents unique challenges. Its multi-modal, multi-temporal, and multi-resolution characteristics—spanning optical, SAR, and multi-spectral imagery—demand expert-level interpretation that MLLMs struggle to achieve. Moreover, most existing MLLMs are designed for instruction-following or descriptive tasks~\cite{sun2021pbnet,li2021deep,wang2022unetformer,zhu2022land}, and thus fail to deliver the analytical reasoning required for geospatial interpretation. These limitations call for domain-aware intelligent systems that understand user intent and autonomously orchestrate specialized models.

AI Agents~\cite{xi2025rise}, autonomous systems that interpret user intent, plan multi-step tasks, invoke tools, and refine decisions from intermediate outcomes~\cite{reed2022generalist,park2023generative,wang2024survey}, offer a promising pathway for remote sensing intelligence. 
Recent advancements in agent frameworks—often powered by large language models—have demonstrated strong potential in coordinating multiple tools, decomposing complex tasks, and flexibly adapting to diverse user goals through modular design~\cite{ruan2023tptu,xi2025rise}, with successful adoption in domains such as next-generation wireless networks~\cite{cui2025overview,chen2025towards}.

Although a few works in the remote sensing field use the term ``agent'' in their paper~\cite{du2023tree,liu2024change,zhu2024rs}, most of them are designed for narrowly scoped tasks and lack a unified architecture that can generalize across heterogeneous remote sensing applications. A representative example is RS-ChatGPT~\cite{guo2024remote}, which integrates ChatGPT~\cite{brown2020language} with a collection of pre-trained remote sensing networks to perform tasks such as object detection, scene classification, and image captioning. 
However, RS-ChatGPT encounters difficulties when scaling to more diverse tools or addressing problems that require structured procedural reasoning and specialized domain knowledge, revealing the absence of a unified, domain-aware agent framework for real-world remote sensing analysis. 

In this paper, we propose RS-Agent, a domain-specific AI agent designed to bridge this gap by integrating LLM reasoning with expert knowledge for professional remote sensing applications.
As shown in Fig.~\ref{fig: teaser figure}, RS-Agent comprises four components that mirror the typical remote sensing workflow: \textbf{(i)} a LLM-based Central Controller that interprets queries, plans tasks, executes tools, and aggregates results; \textbf{(ii)} a Toolkit of SOTA methods covering tasks such as denoising, super-resolution, detection, captioning, and scene classification (Fig.~\ref{fig:qualitative_result}); \textbf{(iii)} a Solution Space that provides task-specific solutions to guide tool selection; \textbf{(iv)} a Knowledge Space that supplies domain expertise for remote sensing tasks.

The core functionality of RS-Agent lies in interpreting user instructions and converting them into executable workflows. To this end, we design a structured agent architecture that combines modular prompting with dynamic context tracking, enabling the Central Controller to plan workflows adaptively from current inputs, tool outputs, and contextual memory~\cite{wang2024robust,lv2024theoretical}.
In the Solution Space, to enable RS-Agent to reason like a remote sensing analyst, we build a Solution Database storing professional knowledge for tool use, and propose Task-Aware Retrieval, which infers task types from user queries and retrieves expert-defined solution procedures for accurate tool selection and step-by-step decomposition.
To address the domain knowledge limitations of general-purpose models~\cite{brown2020language,touvron2023llama,bai2023qwen}, we develop a dedicated Knowledge Database, tailored to the remote sensing domain. Building on this, we propose DualRAG, which combines weighted keyword-aware retrieval with global semantic search to enable the agent to retrieve more relevant and accurate domain-specific knowledge. 

We conduct extensive experiments across 9 datasets and 18 remote sensing tasks. RS-Agent achieves over 95\% task planning accuracy, substantially outperforming the SOTA baseline RS-ChatGPT~\cite{guo2024remote}, and consistently surpasses single-model performance on scene classification, RS visual question answering, and object counting. The proposed DualRAG further improves domain knowledge retrieval. RS-Agent is LLM-agnostic; we validate its generality with ChatGPT~\cite{brown2020language}, LLaMA~\cite{touvron2023llama}, Qwen~\cite{yang2024qwen2}, and DeepSeek~\cite{guo2025deepseek}.

The contributions of this paper are summarized as follows:
\begin{itemize}
\item \textbf{RS-Agent: A Comprehensive Agent Framework For Remote Sensing.} We present RS-Agent, an architecture designed for remote sensing scenarios that interprets user queries and orchestrates diverse tools for accurate task execution. Its four components—Central Controller, Toolkit, Solution Space, and Knowledge Space—work in concert to enable robust performance across applications.

\item \textbf{Domain-Adapted Task Planning Solution Space with Task-Aware Retrieval method.} To enhance the agent's task planning accuracy, we propose a domain-adapted Task-Aware Retrieval method. It exploits the structured nature of remote sensing workflows by first inferring a task type and then retrieving expert-curated solution templates as a planning prior. Leveraging domain-specific task knowledge, RS-Agent emulates the decision-making and tool selection processes of professional remote sensing analysts for complex and multi-step tasks.

\item \textbf{Knowledge Space with DualRAG to Retrieve Professional Domain Knowledge.} To strengthen RS-Agent's remote sensing domain knowledge, we propose DualRAG, a domain-adapted retrieval augmented generation method that assigns weights to extracted keywords and performs dual path retrieval, thereby enhancing the accuracy and relevance of knowledge retrieval for geospatial analysis tasks. Its dual-path design addresses the heterogeneous nature of remote sensing queries that require both macro-level context and fine-grained technical details.

\item \textbf{Superior Performance on Various Applications.} Extensive experiments demonstrate that RS-Agent consistently surpasses prior SOTA MLLMs across remote sensing applications and substantially boosts task planning accuracy, marking a step forward in adapting AI agents to remote sensing.

\end{itemize}

\section{Related Works}\label{sec:Related Works}
\subsection{Remote Sensing Multimodal Large Language Models }

MLLMs have transformed human-AI interaction in earth observation by aligning visual and semantic features for natural-language interpretation of remote sensing imagery. Early work focused on captioning~\cite{shi2017can,wang2020word} and VQA~\cite{lobry2020rsvqa}, but lacked flexibility for open-ended queries, motivating general-purpose models. RSGPT~\cite{hu2025rsgpt} and GeoChat~\cite{kuckreja2023geochat} adapted LLaVA to remote sensing, supporting multiturn dialogue and visual geo-localization, while LHRS-Bot~\cite{muhtar2024lhrs} introduced a multi-level vision-language alignment for SOTA perception. EarthVQANet~\cite{wang2024earthvqanet} and SkySenseGPT~\cite{luo2024skysensegpt} added multi-task learning and knowledge-graph integration to enhance reasoning. For temporal dynamics, ChangeCLIP~\cite{dong2024changeclip} and SkyEyeGPT~\cite{zhan2025skyeyegpt} extended to change captioning and video understanding, and UniRS~\cite{li2024unirs} unified single-frame, bi-temporal, and video inputs.

However, the end-to-end paradigm has intrinsic limitations: current MLLMs struggle with deep reasoning, often hallucinate on quantitative tasks such as object counting, and rely on static knowledge that requires prohibitive retraining to update~\cite{kuckreja2023geochat}. RS-Agent addresses these issues by shifting to agentic orchestration—leveraging LLM planning to invoke specialized tools and retrieve dynamic knowledge without retraining.

\subsection{LLM-based Agents in Remote Sensing}

LLM-based agents extend static MLLMs with autonomous planning and tool orchestration, perceiving environments and refining decisions via feedback loops~\cite{wu2023autogen,park2023generative}. In remote sensing, such agents have evolved from single-task specialists to general-purpose frameworks.
Early works targeted vertical applications: TreeGPT~\cite{du2023tree} tailors an expert system for forestry with tree segmentation and ecological parameter estimation, while Change-Agent~\cite{liu2024change} enables interactive change detection and captioning. Moving toward broader applicability, RS-ChatGPT~\cite{guo2024remote} pioneered tool chaining by linking ChatGPT with discriminative models for tasks like detection and classification. More recently, ThinkGeo~\cite{shabbir2025thinkgeo} and EarthAgent~\cite{feng2025earthagent} established comprehensive benchmarks for assessing agentic capabilities.

However, a structural gap remains: most baselines rely on generic agent paradigms (e.g., standard ReAct loops) that lack cognitive structures for remote sensing's procedural complexity, and depend on the LLM's static parametric memory rather than dynamic professional knowledge. RS-Agent bridges these gaps with a Solution Space for expert-level procedural guidance and a DualRAG mechanism for structured knowledge retrieval.

\subsection{Retrieval-Augmented Generation}
Retrieval-Augmented Generation (RAG) addresses LLMs' lack of domain-specific training data~\cite{kenneweg2024retrieval}. Standard dense or sparse retrieval~\cite{gao2023precise,gao2023retrieval,chan2024rq} works for general queries but falters on multi-hop or global-aggregation tasks~\cite{tang2024multihop}. Graph-RAG approaches~\cite{edge2024local,zhu2025knowledge} integrate entity-relation structures; notably, LightRAG~\cite{guo2024lightrag} combines graph indexing with keyword-guided queries, but builds a single retrieval vector by concatenating all keywords with equal weight. In remote sensing, where queries mix critical technical specifications (e.g., sensor types or resolutions) with generic terms, this uniform concatenation dilutes key constraints. We therefore propose DualRAG, which combines holistic query-based and decomposed keyword-based retrieval with a dynamic weight allocation mechanism.

\begin{figure*}[t!]
\begin{center}
\includegraphics[width=\linewidth]{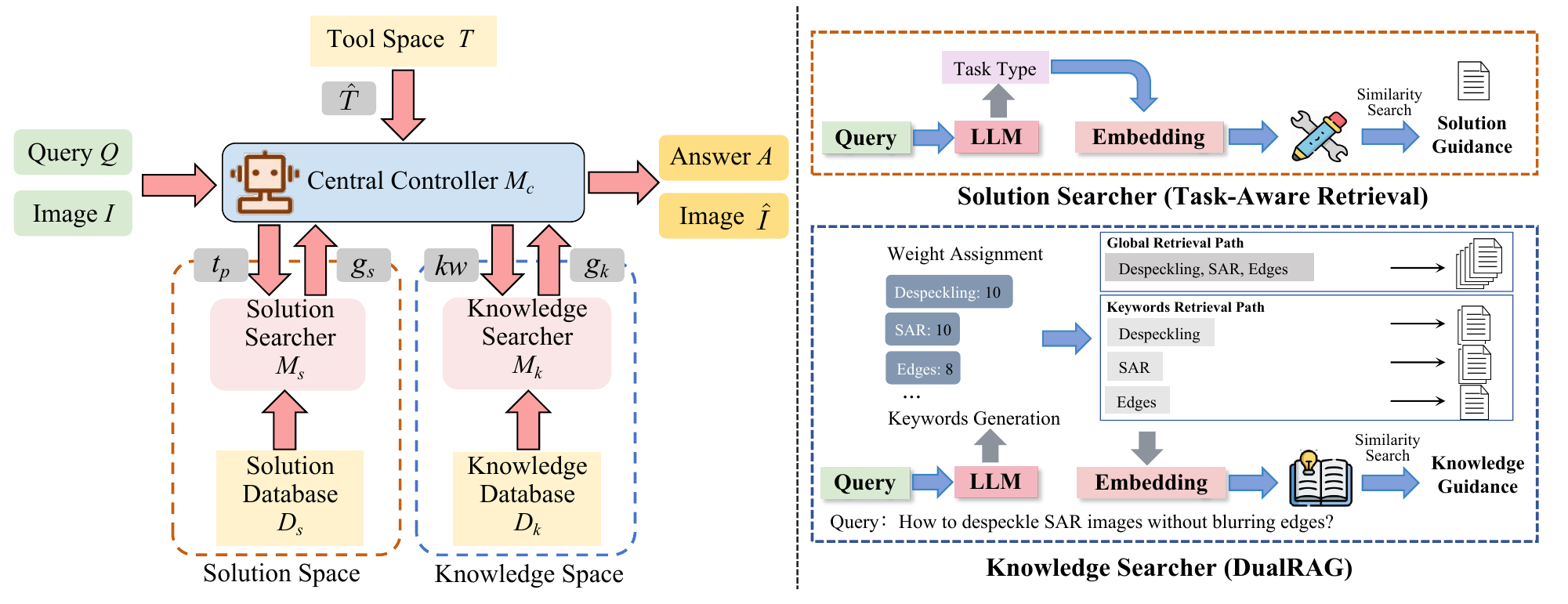}
\caption{Framework of RS-Agent. The left side shows the overall architecture, including key components like the Central Controller, Solution Space, Knowledge Space, and Toolkit. The right side details the Solution Searcher $M_s$ and Knowledge Searcher $M_k$. The $M_s$ uses the Task-Aware Retrieval method for solution guidance $g_s$ generation. The $M_k$ adopts a DualRAG strategy to retrieve relevant documents. }

\label{fig:method}
\vspace{-1.6em}
\end{center}
\end{figure*}

\section{Remote Sensing Agent~(RS-Agent)}

In this section, we formally define the problem addressed by RS-Agent and detail its architectural design. We then elaborate on the four core components that enable robust geospatial reasoning.

\subsection{Overall Architecture}

\subsubsection{Problem Formulation}
RS-Agent serves as an intelligent interface that maps a user query $Q$ and a remote sensing image $I$ to a textual answer $A$ and, optionally, a processed visual output $\hat{I}$. Unlike general vision-language tasks, remote sensing queries are often \textit{under-specified} (e.g., ``Assess the damage'' implies change detection, area calculation, and severity grading) and \textit{domain-dependent} (requiring sensor knowledge), so standard MLLMs $f(Q, I)$ fail to capture the implicit procedural requirements.
RS-Agent therefore formulates the problem as multi-stage reasoning, introducing three latent variables: task type $t_p$ (analytical objective), solution guidance $g_s$ (expert procedural workflows), and domain knowledge $g_k$ (geospatial context and specifications), which together bridge user intent and tool execution.

\subsubsection{Architectural Design}

To address procedural ambiguity and knowledge sparsity in remote sensing queries, RS-Agent adopts a task-centric architecture (Fig.~\ref{fig:method}). Rather than mapping $Q$ directly to tool actions, it first abstracts user intent into explicit task types; this Task Inference step decouples intent understanding from execution, enabling retrieval of expert-verified workflows from the Solution Space and geospatial context from the Knowledge Space before tool invocation. RS-Agent comprises four core components:

(1) \textbf{Central Controller ($M_c$)}: Serves as the decision-making core of the agent, built upon a Large Language Model. It is responsible for four key functions: (i) \textbf{Perception}: Analyzing user instruction $Q$ and image metadata $I$; (ii) \textbf{Planning}: Consulting the Solution Space to decompose abstract queries into executable workflows; (iii) \textbf{Retrieval}: Querying the Knowledge Space for domain support when parameter or context ambiguity arises; and (iv) \textbf{Action}: Dynamically orchestrating tools and synthesizing the final response $A$ based on execution results.

(2) \textbf{Toolkit ($T$)}: A modular and extensible collection of state-of-the-art remote sensing tools covering diverse modalities (e.g., optical imagery, SAR) and tasks spanning multiple analysis levels (e.g., low-level processing tasks such as denoising and super-resolution, and high-level semantic tasks such as detection, segmentation, classification, and change analysis). Unlike end-to-end MLLMs where tool capabilities are frozen in model weights, the Toolkit design is modular and extensible, supporting plug-and-play integration. It allows for the seamless integration of new SOTA models or specialized tools without requiring the retraining of the Central Controller, thereby ensuring the agent remains at the forefront of algorithmic performance.

(3) \textbf{Solution Space ($D_s$)}: A repository of expert-verified standard operating procedures addressing the procedural ambiguity of remote sensing tasks. A \textbf{Solution Searcher} ($M_s$) with a \textbf{Task-Aware Retrieval} mechanism conditions retrieval on the inferred task type ($t_p$) rather than the raw query, fetching solution guidance ($g_s$) that encodes professional tool execution sequences and ensures rigorous workflows over zero-shot planning.

(4) \textbf{Knowledge Space}: Complementing the Solution Space, the Knowledge Space stores structured remote sensing knowledge, such as definitions, operational principles, or region-specific metadata. During task planning, the Controller queries this space to resolve ambiguities or retrieve explanatory information. 
To ensure relevant knowledge access, this module features a \textbf{Knowledge Searcher} ($M_k$) driven by our proposed \textbf{DualRAG} strategy, which combines global retrieval and keyword-guided retrieval to capture both high-level contextual knowledge and task-relevant details.

\begin{algorithm}[tb]
\footnotesize
\caption{Task-Centric Workflow of RS-Agent}
\label{alg:Workflow}
\begin{algorithmic}[1]

\REQUIRE User query $Q$ and input remote sensing image $I$
\ENSURE Final answer $A$ and processed image $\hat{I}$ (if required)

\STATE \textbf{Initialization}
\STATE Initialize the central controller $M_c$ with $(Q, I)$

\STATE \textbf{Task Inference}
\STATE $M_c$ infers and decomposes the latent task representation $tp$ from $(Q, I)$,
optionally grounded by retrieved domain knowledge

\STATE \textbf{Task-to-Solution Grounding}
\STATE $M_c$ grounds $tp$ to task-specific solution guidance $g_s$
retrieved from the Solution Space $D_s$ via the Solution Searcher $M_s$

\STATE \textbf{Pipeline Orchestration}
\STATE Based on $(Q, tp, g_s)$, $M_c$ constructs an ordered tool execution plan
$\hat{T} = \{T_1, T_2, \dots\}$

\STATE \textbf{Tool Execution}
\STATE Execute tools in $\hat{T}$ to process the input image $I$ and intermediate results,
producing outputs and the processed image $\hat{I}$ when applicable

\STATE \textbf{Knowledge-Grounded Interpretation (if required)}
\STATE $M_c$ enriches outputs with domain knowledge retrieved via $M_k$ from $D_k$

\STATE \textbf{Response Generation}
\STATE Generate final answer $A$

\end{algorithmic}
\end{algorithm}

Algorithm~\ref{alg:Workflow} summarizes the end-to-end workflow in six stages:

\textbf{(1) Initialization:} $M_c$ receives the user query $Q$ (e.g., ``How many airplanes are parked in this image, and what are their categories?'') and image $I$.

\textbf{(2) Task Inference:} $M_c$ analyzes $Q$ to infer the underlying task type(s) $tp$ by mapping the query to predefined remote sensing task categories. When the query intent is ambiguous or under-specified, $M_c$ may optionally retrieve domain knowledge from $D_k$ to guide accurate task classification.

\textbf{(3) Solution Retrieval:} The inferred task type $tp$ is passed to the Solution Searcher $M_s$, which retrieves solution guidance $g_s$ from the Solution Database $D_s$. This guidance encodes task-specific procedural workflows—for instance, performing object detection before fine-grained category classification for the aircraft counting query.

\textbf{(4) Pipeline Orchestration:} Conditioned on $(Q, g_s)$, $M_c$ constructs an executable tool chain $\hat{T} = \{T_1, T_2, \dots\}$ from the Toolkit $T$.

\textbf{(5) Tool Execution:} $M_c$ invokes the tools in $\hat{T}$ sequentially or in parallel as required, processing the input image $I$ and intermediate results. This stage produces task-specific outputs and, when applicable, a processed visual result $\hat{I}$ (e.g., detection bounding boxes).

\textbf{(6) Knowledge-Grounded Response Generation:} If specialized knowledge is needed, $M_c$ invokes $M_k$ to retrieve $g_k$ from $D_k$ via DualRAG, then integrates tool outputs and knowledge into the final answer $A$.

\vspace{-0.48em}
\subsection{Central Controller}
\vspace{-0.6em}

The Central Controller $M_c$ is the brain of RS-Agent, managing multi-step reasoning and tool execution through modular interfaces. Built on an LLM with LangChain~\cite{langchain}, $M_c$ supports task decomposition, context management, and dynamic tool invocation, and adopts a two-stage prompt-based reasoning process that embeds remote-sensing domain knowledge to guide task formulation.

\textbf{Stage 1. Task Inference and Solution Retrieval.} 

Given a user query $Q$, $M_c$ infers task type(s) $tp$ by aligning $Q$ with a predefined set of supported remote-sensing tasks via a structured classification prompt. Since user queries often describe high-level goals without specifying task formulation, scale, or temporal setting, the prompt guides $M_c$ to distill remote-sensing attributes—sensor/modality, spatial granularity, temporal configuration, and analytical operation—and produce a compact, machine-readable task representation drawn exclusively from the supported task set.

The inferred $tp$ is then forwarded to the Solution Searcher $M_s$, which retrieves candidate solution $g_s$—an expert-defined guideline for solving the task.

\textbf{Stage 2. Task Planning and Tool Execution.} Leveraging the retrieved $g_s$ and the original $Q$, the Central Controller $M_c$ constructs a detailed task plan by selecting the most appropriate tools and arranging them into a logical, step-by-step workflow tailored to the user's intent. To facilitate the memory mechanism, $M_c$ incorporates prior dialogue history to ensure accurate, context-aware reasoning.

\subsection{Toolkit}
The Toolkit $T$ is the action engine of RS-Agent, providing a structured and extensible suite of analysis tools. Given remote sensing's strong modality dependency, multi-scale nature, and geometry–semantics coupling, $T$ adopts a three-level hierarchy (Fig.~\ref{fig:graph_toolkit}(b)): first, by sensing modality (optical vs.\ SAR); second, by analysis level (low-level processing such as denoising and super-resolution; high-level semantics such as classification and interpretation); third, by geometric and semantic characteristics including target semantics, spatial scale, resolution, and analysis granularity. All tools are accessed via standardized APIs for modular orchestration, and the Toolkit is extensible to new tools as needed. The framework currently includes 27 tools, with a representative subset shown in Fig.~\ref{fig:qualitative_result}. A complete list of all supported tools is provided in Table~S3 in the Supplementary File.

\subsection{Solution Space}\label{sec:solution_space}

The Solution Space supports accurate task planning by compensating for LLMs' lack of domain-specific solution strategies, especially when task instructions are ambiguous, underspecified, or implicitly formulated~\cite{qin2023tool,shapira2023clever}—scenarios common in remote sensing. It comprises two components: the Solution Database $D_s$, storing expert-curated solutions, and the Solution Searcher $M_s$, which retrieves the most appropriate solution for a given user query.

\subsubsection{Solution Database}
To provide RS-Agent with a reliable source of task-solving knowledge, we construct a curated Solution Database $D_s = \{ d_s^1, d_s^2, \dots \}$, which stores expert-defined solutions for commonly encountered remote sensing tasks. This database acts as a structured knowledge base that guides task planning and tool invocation.

Solutions in $D_s$ are built in three stages. First, domain experts manually design core solutions based on established analysis paradigms and tool capabilities, encoding task formulations, recommended tool chains, input requirements, and expected outputs. Second, this set is expanded via instruction-based LLM generation to produce paraphrased variants that improve retrieval robustness. Third, all generated variants are verified for correctness and executability by executing the prescribed tool chain on representative test inputs and checking that input/output formats are consistent across the pipeline.

Each solution document follows a unified structure: task description, applicable conditions, recommended tool sequence, input formats, and expected outputs. For example, a solution for object counting specifies an object-level detection tool followed by a counting step, translating a high-level request into an executable workflow.
In the current implementation, $D_s$ contains 34 solution documents in total: 27 single-tool templates and 7 multi-tool workflow templates. The 27 single-tool templates correspond one-to-one with the 27 tools in the Toolkit (Fig.~\ref{fig:graph_toolkit} (b) and Table~S3 in Supplementary File), each encoding the recommended invocation procedure for one tool. The 7 multi-tool templates are designed for RSVQA tasks, encoding composite pipelines that chain image preprocessing tools (e.g., denoising, super-resolution) with visual question answering.

\subsubsection{Solution Searcher with Task-Aware Retrieval}

To enable RS-Agent to accurately locate the most relevant solution for a given task, we design the Solution Searcher, a module dedicated to retrieving appropriate entries from the Solution Database. To enhance its effectiveness, we propose a Task-Aware Retrieval method, which augments the capabilities of large language models by integrating Retrieval-Augmented Generation (RAG) into the solution selection process. This design leverages remote sensing's structured task taxonomy. Since user queries describe goals without explicit task specifications, Task-Aware Retrieval first infers the task type, then uses it as the retrieval key for more effective matching.

As shown in the top right of Fig.~\ref{fig:method}, Task-Aware Retrieval has two steps: Task Inference, which identifies the task type from the query, and Solution Retrieval, which uses this type to fetch the most relevant solution.

\textbf{Task Inference:} When a user query $Q$ is received, the Central Controller $M_c$ first interprets the user’s intent and infers the underlying $tp$ using a system prompt designed to guide task recognition. This step can be viewed as a query rewriting process, where the LLM reformulates or abstracts the input into a semantically meaningful task label, enabling more structured downstream retrieval. Critically, this prompt encodes remote sensing domain knowledge---including sensing modality distinctions, analysis-level hierarchies, and task granularity cues---enabling the LLM to resolve ambiguities that general-purpose task inference would miss. The complete prompt design is provided in Supplementary File~A.1.

\textbf{Solution Retrieval:} The inferred $tp$ is then passed to the Solution Searcher $M_s$, which retrieves relevant strategies from $D_s$ to support tool selection and reasoning. 
To achieve retrieval, each solution document in $D_s$ and $tp$ are encoded into dense vector representations using a sentence embedding model~\cite{chen2024bge}. 
$M_s$ then employs FAISS~\cite{jegou2022faiss} as a high-performance vector similarity search function $f_s$ to retrieve relevant documents and form the Solution Guidance $g_s$:
\begin{equation}
g_s = f_s(tp, D_s, k), \quad k = |tp| \,,
\end{equation}
where $|tp|$ is the number of task types in $tp$. This yields a compact set of complementary solution documents covering the implied task requirements.

By combining retrieval and generation, Task-Aware Retrieval enables RS-Agent to leverage expert-defined tool solutions and reliably handle complex remote sensing tasks.

\vspace{-0.8em}
\subsection{Knowledge Space}
\vspace{-0.5em}
The Knowledge Space strengthens RS-Agent's ability to handle queries requiring specialized domain knowledge. It contains two components: a Knowledge Database $D_k$ storing curated remote sensing knowledge, and a Knowledge Searcher $M_k$ that retrieves relevant documents from $D_k$. 
\vspace{-0.5em}
\subsubsection{Knowledge Database}
\begin{figure*}[t]
  \centering
  \includegraphics[width=\linewidth]{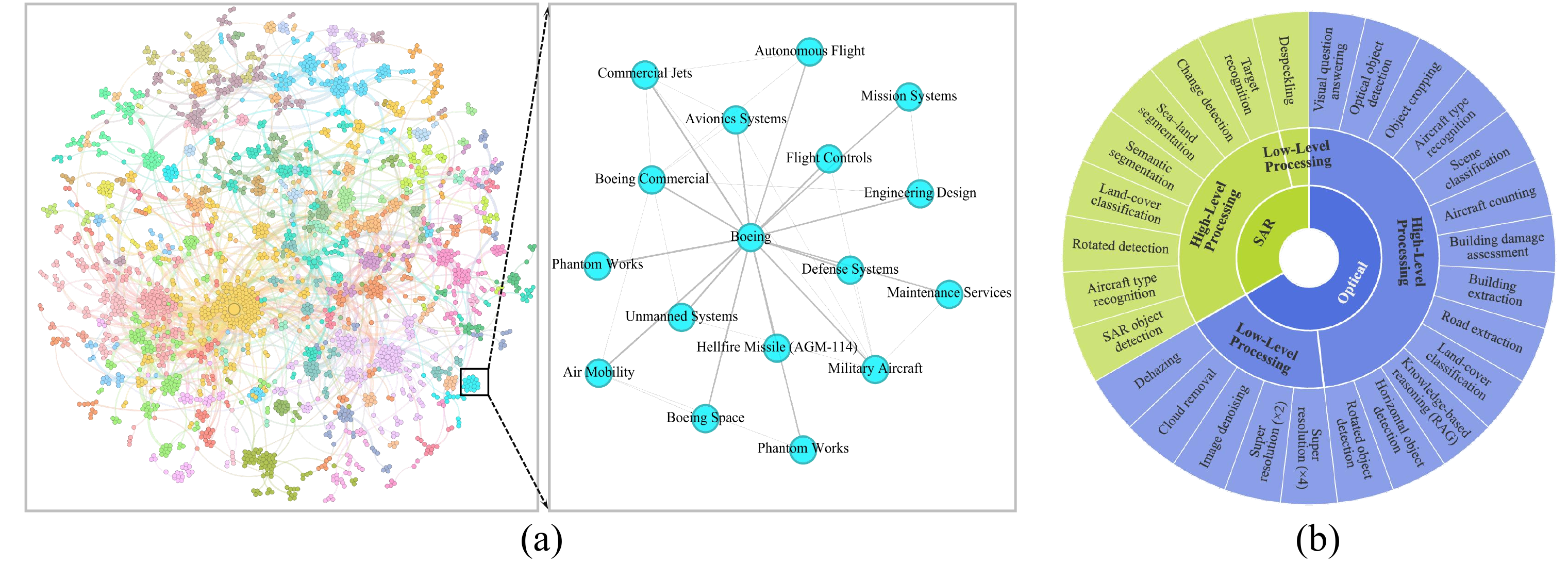}
  \caption{Overview of the RS-Agent knowledge representation and toolkit organization.(a) RSaircraft knowledge graph. Nodes denote remote sensing knowledge entities, and edges represent typed semantic relations, forming a dense core with sparse peripheral components.(b) Hierarchical structure of the Toolkit $T$. Tools are organized into a three-level hierarchical structure and are accessed via standardized APIs for modular orchestration.}
    
  \label{fig:graph_toolkit}
  \vspace{-1.0em}
\end{figure*}

The Knowledge Database $D_k$ serves as a structured repository of domain knowledge that supports RS-Agent in tasks requiring factual accuracy, technical terminology, or background-specific reasoning. 
To validate RS-Agent’s capability in retrieving and generating domain-specific knowledge, $D_k$ is designed as a general-purpose remote sensing knowledge corpus, covering concepts related to sensing modalities, platforms, targets, observation principles, and domain-specific semantics. The knowledge corpus is collected from Wikipedia and preprocessed through cleaning and normalization to form semantically coherent text chunks suitable for structured knowledge construction. To evaluate RS-Agent’s capability in retrieving and generating domain-specific knowledge, we further instantiate a domain-focused subset of $D_k$ as a case study. Specifically, as the current RS-Agent is initially developed to support aircraft-related remote sensing analysis, we construct a specialized knowledge subset, denoted as \textit{RSaircraft}, which contains curated information on 65 aircraft categories, including fighter jets, commercial airliners, drones, and helicopters.
This subset exemplifies how the general knowledge corpus can be adapted to other specific application domains. The Knowledge Database is designed for extensibility. Its knowledge graph storage represents content as entity--relation triples in vector spaces, decoupling the schema from any sub-domain ontology. The DualRAG retrieval logic is likewise decoupled from the stored content, so switching corpora requires no pipeline changes. A standardized API between the Knowledge Space and the Central Controller further ensures that new corpora can be ingested without modifying the agent logic. Consequently, expanding to new sub-domains requires only corpus ingestion and knowledge graph reconstruction. We adopt a knowledge graph-based storage structure~\cite{guo2024lightrag}, in which each text chunk is processed by an LLM to extract entities and relational triples. These entities and relations are embedded into separate vector spaces, enabling structure-aware semantic retrieval. This representation supports multi-hop reasoning and domain-informed query expansion, facilitating more accurate and context-rich responses. A local subgraph of the constructed knowledge graph is illustrated in Figure~\ref{fig:graph_toolkit} (a).

\subsubsection{Knowledge Searcher with DualRAG}

The Knowledge Searcher retrieves relevant documents from $D_k$ for domain-specific queries. Existing single-query or keyword-based methods struggle with complex remote sensing queries, where entities, relations, and constraints are implicit, unevenly emphasized, or scattered across fragments. We thus propose \emph{DualRAG}, a domain-adapted retrieval method tailored for remote sensing knowledge retrieval, which introduces weighted keyword-aware query decomposition and a dual-path retrieval architecture.

As shown in Fig.~\ref{fig:method}, DualRAG has two components: (1) Keyword Generation and Weight Assignment, which scores semantically important keywords to emphasize key entities and relations; and (2) Dual-Path Retrieval, which fuses global semantic search with weighted keyword-level retrieval.

\textbf{Keyword Generation and Weight Assignment:}
Given a user query $Q$, the Knowledge Searcher $M_k$ generates a ranked set of retrieval-oriented keywords and assigns an importance weight to each keyword based on prompt-defined prioritization and scoring rules to facilitate remote-sensing knowledge retrieval. In particular, $M_k$ prioritizes RS domain-specific and discriminative terms while filtering out generic or low-information words, producing a compact weighted representation. This weighted keyword set acts as a structured query expansion that narrows the retrieval space and steers the downstream stage toward the most relevant remote-sensing knowledge.
\begin{equation}
\{(kw_i, w_i)\}_{i=1}^m = M_c(Q), \quad \text{where } w_i \in [1, 10] \,.
\end{equation}

Each tuple $(kw_i, w_i)$ represents a keyword $kw_i$ and its corresponding importance weight $w_i$, where larger $w_i$ indicates higher expected utility for retrieval. The weighting is governed by a domain-informed prioritization hierarchy that reflects the structure of remote sensing queries, so that retrieval-critical aspects are emphasized. The complete keyword extraction prompt is provided in Supplementary File~A.2. Here, $m$ denotes the total number of extracted keyword--weight pairs.

\textbf{Dual-Path Retrieval:}
To enhance the relevance and coverage of the retrieved knowledge documents, our DualRAG adopts a dual-path retrieval strategy that combines global and keyword-level search. In the global retrieval path, the generated $kw_i$ are concatenated to form a single string: 
\begin{equation} kw_{\text{global}} = \text{Concat}(kw_1, kw_2, \dots) \,.
\end{equation} 
This string retrieves the top-$N$ documents from the Knowledge Database $D_k$ via similarity search algorithm $f_k$: 
\begin{equation} 
d_{\text{global}} = f_k(kw_{\text{global}}, D_k, N) \,.
\end{equation} 
This path captures the overall semantic structure and inter-keyword dependencies.

In the keyword retrieval path, each $kw_i$ is treated as an individual query. The number of the retrieved knowledge documents $n_i$ is allocated based on the assigned weight $w_i$: 
\begin{equation} 
n_i = \left\lfloor \frac{w_i}{\sum_{j=1}^{m} w_j} \times N \right\rfloor \,,
\end{equation}
where each $n_i$ is obtained by floor division so that keywords with higher importance weight are allocated more documents. The remaining quota $N - \sum_i n_i$ is distributed one-by-one to keywords in descending order of $w_i$, ensuring $\sum_i n_i = N$. Then, each $kw_i$ retrieves its own top-$n_i$ documents: \begin{equation} 
d_{\text{keyword}} = \bigcup_{i=1}^{m} f_k(kw_i, D_k, n_i) \,,
\end{equation} 
where $m$ is the number of keywords. This path retrieves fine-grained information complementing the global path.

The retrieved documents form the Knowledge Guidance \( g_k = d_{\text{global}} \cup d_{\text{keyword}} \), which provides the LLM with domain-specific knowledge. Duplicate documents appearing in both paths are removed during the union operation, and each document is retained at most once in $g_k$.

By combining both paths, DualRAG balances semantic coherence and informational diversity. Unlike LightRAG~\cite{guo2024lightrag}'s single pipeline, DualRAG's dual-path structure separates global and local retrieval scopes—capturing both macro-level semantics and fine-grained entity details—and is particularly effective for complex remote sensing queries involving multiple entities.

\section{Experiments}

In this section, we present comprehensive experiments to validate the effectiveness of RS-Agent across multiple dimensions. We organize our evaluation as follows: (\textit{i}) implementation details (Section 4.1); (\textit{ii}) task planning accuracy versus SOTA agents and across open-source and proprietary LLMs (Section 4.2); (\textit{iii}) performance over single-model baselines on scene classification, visual question answering, and object counting (Section 4.3); (\textit{iv}) ablation studies on Task-Aware Retrieval and DualRAG (Sections 4.4 and 4.5); (\textit{v}) qualitative results on multi-step reasoning (Section 4.6).

\subsection{Implementation Details}

To ensure the flexibility and generalizability of RS-Agent, we build the Central Controller with the LangChain framework~\cite{langchain} and general-purpose large language models. Unless otherwise specified, RS-Agent uses the Qwen 2.5 32B-Instruct model~\cite{yang2024qwen2}, balancing the model capability with inference speed. 

For Task-Aware Retrieval, we employ the open-source m3e-base~\cite{chen2024bge} embedding model and index vectors with FAISS~\cite{jegou2022faiss}. For DualRAG, we return the top-$N$ documents with $N{=}30$.

RS-Agent is equipped with a toolkit of specialized tools designed to support diverse remote sensing tasks, ranging from low-level vision operations to high-level semantic tasks. For performance evaluation, we instantiate the toolkit with a representative set of 18 tasks to benchmark RS-Agent under a controlled and reproducible setting.

\begin{table*}[tb]
\centering
\captionsetup{justification=centering, labelsep=space, font=small, textfont=normal, labelfont=bf}
\caption{Comparison of Task Planning Accuracy with SOTA RS-ChatGPT}
\label{tab:tool_accuracy}
\resizebox{\textwidth}{!}{ 
\begin{tabular}{lcccccc}
\toprule
\multirow{2}{*}{Task} & \multicolumn{2}{c}{gpt-3.5-turbo-1106} & \multicolumn{2}{c}{gpt-3.5-turbo} & \multicolumn{2}{c}{gpt-4o-mini} \\
\cmidrule(lr){2-3} \cmidrule(lr){4-5} \cmidrule(lr){6-7}
& RS-ChatGPT & RS-Agent~(Ours) & RS-ChatGPT & RS-Agent~(Ours) & RS-ChatGPT & RS-Agent~(Ours) \\
\midrule
Object Counting & 47.37\% & 94.74\% & 97.74\% & 100\% & 100\% & 100\% \\
Object Detection & 78.95\% & 84.21\% & 52.63\% & 89.47\% & 42.11\% & 94.74\%\\
Land Use Segmentation & 90.48\% & 95.24\% & 95.24\% & 95.24\% & 90.48\% & 90.48\%\\
Image Captioning & 42.86\% & 57.14\% & 85.71\% & 61.90\% & 100\% & 95.24\%\\
Scene Classification & 90.48\% & 85.71\% & 80.95\% & 100\% & 100\% & 90.48\%\\
Instance Segmentation & 57.89\% & 78.95\% & 78.95\% & 68.42\% & 89.47\% & 100\%\\
Edge Detection & 100\% & 100\% & 84.21\% & 100\% & 94.74\% & 100\%\\
\midrule
Average Accuracy & 72.66\% & \textbf{84.89\%} & 82.01\% & \textbf{87.77\%} & 88.49\% & \textbf{95.68\%}\\
\bottomrule
\end{tabular}}
\end{table*}

\begin{table*}[tb]
\centering
\captionsetup{justification=centering, labelsep=space, font=small, textfont=normal, labelfont=bf}
\resizebox{\textwidth}{!}{ 
\begin{threeparttable}

\caption{Task Planning Accuracy of RS-Agent with Different LLMs on 10 Representative Tasks. Complete results for all tasks are reported in Table~S2 (Supplementary File).}
\label{tab:ablation_llm}
\begin{tabular}{lccccccccc}
\toprule
\multirow{2}{*}{Task} & \multicolumn{3}{c}{ChatGPT} & \multicolumn{2}{c}{LLaMa 3.1} & \multicolumn{3}{c}{Qwen2.5} & \multicolumn{1}{c}{DeepSeek} \\
\cmidrule(lr){2-4} \cmidrule(lr){5-6} \cmidrule(lr){7-9} \cmidrule(lr){10-10}
 & \makecell{3.5-turbo-1106 \\ (87.71t/s)} & \makecell{3.5-turbo \\ (65.03t/s)} & \makecell{4o-mini \\ (58.87t/s)} & \makecell{8B \\ (100.78t/s)} & \makecell{70B \\ (17.71t/s)} & \makecell{14B \\ (69.61t/s)} & \makecell{32B \\ (36.77t/s)} & \makecell{72B \\ (16.24t/s)} & \makecell{r1:70B \\ (18.25t/s)} \\
\midrule
Image Captioning     & 55.00\% & 45.00\% & 90.00\% & 15.00\% & 60.00\% & 70.00\% & 80.00\% & 80.00\% & 10.00\% \\
Object Detection    & 75.00\% & 60.00\% & 95.00\% & 30.00\% & 90.00\% & 90.00\% & 85.00\% & 100\% & 85.00\% \\
Scene Classification           & 20.00\% & 90.00\% & 100\% & 80.00\%     & 90.00\%   & 90.00\%   & 100\%   & 100\%   & 50.00\% \\
SAR Detection           & 30.00\% & 100\% & 100\% & 75.00\%     & 95.00\%   & 100\%   & 100\%   & 100\%   & 100\% \\
SAR Plane Classification           & 100\% & 100\% & 100\% & 100\%     & 100\%   & 100\%   & 100\%   & 100\%   & 90.00\% \\
Knowledge Search          & 100\% & 100\% & 100\% & 100\%     & 80.00\%   & 100\%   & 100\%   & 100\%   & 10.00\% \\
Building Extraction           & 10.00\% & 70.00\% & 100\% & 55.00\%     & 100\%   & 100\%   & 100\%   & 100\%   & 100\% \\
Rotated Detection           & 15.00\% & 35.00\% & 100\% & 85.00\%     & 90.00\%   & 100\%   & 100\%   & 100\%   & 100\% \\
Semantic Segmentation           & 60.00\% & 100\% & 100\% & 80.00\%     & 100\%   & 100\%   & 100\%   & 100\%   & 80.00\% \\
Land Use Classification          & 15.00\% & 100\% & 100\% & 75.00\%     & 100\%   & 100\%   & 100\%   & 95.00\%   & 95.00\% \\
\midrule
Average Accuracy         & 48.00\% & 80.00\% & 98.50\% & 69.50\% & 90.50\% & 95.00\% & 96.50\% & 97.50\% & 72.00\% \\
\bottomrule
\end{tabular}
\begin{tablenotes}
\footnotesize
\item[1]``B'' denotes the number of parameters in billions.
\item[2] Numbers in parentheses (t/s) indicate model inference speed in tokens per second on NVIDIA RTX4090 GPUs.
\end{tablenotes}
\end{threeparttable}}
\end{table*}

\begin{table*}[tb]
\centering
\captionsetup{justification=centering, labelsep=space, font=small, textfont=normal, labelfont=bf}
\setlength{\tabcolsep}{3pt}

\begin{tabular}{@{}p{0.44\textwidth}@{\hspace{0.025\textwidth}}p{0.52\textwidth}@{}}

\begin{minipage}[t]{\linewidth}
\centering
\captionof{table}{Object counting accuracy on DOTA~\cite{xia2018dota}.}
\label{tab:count_accuracy}
\resizebox{0.8\linewidth}{!}{
\begin{tabular}{lccc}
\toprule
Method & \makecell{Absolute \\ Accuracy} & \makecell{Interval \\ Match} & \makecell{Relative \\Error} \\
\midrule
GeoChat & 17.65\% & 74.61\% & 0.65\\
LHRS-Bot & 14.92\% & 60.78\% & 0.56\\
\midrule
RS-Agent~(Ours) & \textbf{33.30\%} & \textbf{75.98\%} & \textbf{0.28}\\
\bottomrule
\end{tabular}}
\end{minipage}
&
\multirow[t]{2}{0.52\textwidth}{%
\begin{minipage}[t]{0.52\textwidth}
\vspace*{-0.25em} 
\centering
\captionof{table}{Visual Question Answering Accuracy on RSVQA-LR datasets~\cite{lobry2020rsvqa}. Top: general-purpose MLLMs; bottom: models tailored for remote sensing.}
\label{tab:performance_comparison}
\resizebox{\linewidth}{!}{
\begin{tabular}{lcccc}
\toprule
Method & Rural/Urban & Presence & Compare & Avg. \\
\midrule
LLaVA-1.5~\cite{liu2023improved} & 59.22\% & 73.16\% & 65.19\% & 65.86\% \\
MiniGPTv2~\cite{chen2023minigpt} & 60.02\% & 51.64\% & 67.64\% & 59.77\% \\
InstructBLIP~\cite{dai2024instructblip} & 62.62\% & 48.83\% & 63.92\% & 59.12\% \\
mPLUG-Owl2~\cite{ye2023mplug} & 57.99\% & 74.04\% & 65.04\% & 65.69\% \\
QWen-VL-Chat~\cite{bai2023qwen} & 62.00\% & 47.65\% & 54.64\% & 58.73\% \\
\midrule
RSVQA~\cite{lobry2020rsvqa} & 90.00\% & 87.47\% & 81.50\% & 86.32\% \\
SkyEyeGPT~\cite{zhan2025skyeyegpt} & 88.93\% & 88.63\% & 75.00\% & 84.16\% \\
LHRS-Bot~\cite{muhtar2024lhrs} & 89.07\% & 88.51\% & 90.00\% & 89.19\% \\
GeoChat~\cite{kuckreja2023geochat} & 94.00\% & 91.09\% & 90.33\% & 90.70\% \\
\midrule
RS-Agent~(Ours) & \textbf{97.00\%} & \textbf{91.07\%} & \textbf{90.58\%} & \textbf{90.88\%} \\
\bottomrule
\end{tabular}}
\end{minipage}%
}
\\[1.0em]

\begin{minipage}[t]{\linewidth}
\centering
\captionof{table}{Scene classification accuracy on RSSDIVCS~\cite{li2021learning}, AID~\cite{xia2017aid}, and UCMerced~\cite{yang2010bag}.}
\label{tab:scene_accuracy}
\resizebox{0.8\linewidth}{!}{
\begin{tabular}{lccc}
\toprule
Model & RSSDIVCS & AID & UCMerced\\
\midrule
MiniGPTv2~\cite{chen2023minigpt} & - & 52.60\% & 62.90\%\\
Qwen-VL-Chat~\cite{bai2023qwen} & - & 52.60\% & 62.90\%\\
LLaVA-1.5~\cite{liu2023improved} & - & 51.00\% & 68.00\%\\
GeoChat~\cite{kuckreja2023geochat} & 51.43\% & 72.03\% & 84.43\%\\
LHRS-Bot~\cite{muhtar2024lhrs} & 63.85\% & 91.29\% & 96.63\%\\
\midrule
RS Agent~(Ours) & \textbf{98.00\%} & \textbf{96.88\%} & \textbf{98.63\%}\\
\bottomrule
\end{tabular}}
\end{minipage}
& \\

\end{tabular}
\end{table*}

\subsection{Task Planning Accuracy}
A core capability of RS-Agent is its ability to accurately plan tasks and invoke the appropriate tools in response to user queries. In this section, we evaluate the task planning accuracy of RS-Agent.

\subsubsection{Task Planning Accuracy Compared to SOTA Agent}

\textbf{Dataset.} We construct a dataset for task planning, measuring accuracy by whether the first invoked tool matches the correct one. For fair comparison with RS-ChatGPT~\cite{guo2024remote}, we adopt its seven core tasks: scene classification, land use segmentation, object detection, image captioning, edge detection, object counting, and instance segmentation. We curate $\sim$20 query-solution pairs per task, mixing clear and ambiguous instructions, and use the same backbone LLMs as RS-ChatGPT.

\textbf{Results:} As shown in Table~\ref{tab:tool_accuracy}, RS-Agent consistently outperforms RS-ChatGPT across all backbone LLMs. With gpt-3.5-turbo-1106, accuracy rises from 72.66\% to 84.89\% (+12.23\%); similar gains hold for gpt-3.5-turbo (+5.76\%) and gpt-4o-mini (+7.19\%). These results demonstrate that RS-Agent exhibits superior task comprehension and decision-making ability, allowing it to reliably select appropriate tools and execute tasks more effectively.

\subsubsection{Task Planning Accuracy with Different LLMs}

We evaluate RS-Agent's adaptability when paired with closed-source (GPT series) and open-source LLMs.

\textbf{Datasets.} To comprehensively evaluate RS-Agent's task planning accuracy across different LLM configurations, we construct a new dataset tailored to the full toolkit currently supported by our system. Specifically, we define a set of 18 distinct tasks and design 20 query-solution pairs for each, resulting in a total of 360 evaluation instances. These queries span a diverse range of task types and complexity levels, and include both well-specified and ambiguous instructions to assess the model’s robustness in understanding intent and selecting the appropriate tool. Unlike the previous dataset, which was based on the RS-ChatGPT toolkit for comparative evaluation, this dataset fully leverages RS-Agent's expanded capabilities and provides a more comprehensive assessment of its performance under different LLM configurations.

\textbf{Results.}
Table~\ref{tab:ablation_llm} reports results on 10 representative tasks selected to cover optical, SAR, and cross-modal categories (complete 18-task results are in Table~S2 of Supplementary File). RS-Agent achieves robust tool selection across proprietary and open-source LLMs. Open-source models such as Qwen2.5-72B (97.50\%) and 32B (96.50\%) match or exceed the ChatGPT series. This result highlights that well-tuned open-source models can match the reasoning capabilities of state-of-the-art commercial models in task planning scenarios.

Within a model family, larger parameters generally improve accuracy: Qwen2.5 rises from 14B (95.00\%) to 72B (97.50\%), and LLaMa 3.1 from 8B (69.50\%) to 70B (90.50\%); but inference speed drops accordingly (Qwen2.5-72B: 16.24 t/s vs.\ 14B: 69.61 t/s). Balancing both, Qwen2.5-14B and 32B are practical choices, while LLaMa 3.1-8B underperforms (69.50\%), indicating smaller variants may not generalize to tool planning.

In summary, from the above results, we can observe that the model performance varies significantly across different tasks, especially those with unclear tool boundaries. This highlights the need for clearer task definitions and more robust planning strategies in future work.

\subsection{Performance on Remote Sensing Tasks}

Like MLLMs, agents perform diverse tasks through natural-language interaction. To highlight RS-Agent's advantages, we evaluate it on three representative remote-sensing applications: object counting, visual question answering (VQA), and scene classification.

\subsubsection{Object Counting} Object counting estimates the number of target objects in an image, supporting remote-sensing applications such as vehicle counting for traffic analysis and infrastructure density assessment for urban planning, where accurate quantification is critical for resource management. We employ YOLOv8x-obb~\cite{yolov8_github}, pre-trained on DOTAv1~\cite{xia2018dota}, which excels at oriented object detection.

\textbf{Datasets.} We use the DOTAv1~\cite{xia2018dota} validation set (458 images, 1{,}099 questions) with the prompt: ``What is the number of the \{object\}? Answer the question using a number.'', where \{object\} is one of 15 typical remote-sensing target types (e.g., plane, ship, bridge).

\textbf{Metric.} We use three metrics: \emph{absolute accuracy} (percentage of predictions exactly matching the ground truth), \emph{interval-matching accuracy} (a prediction is correct if it falls in the same predefined range as the ground truth, with intervals 0, 1--10, 11--100, 101--1000, and $>$1000), and \emph{relative error} $e_r$, defined as
\begin{equation}
   e_r = \frac{1}{N} \sum_{i=1}^{N} \log \left( 1 + \frac{|{gt}_i - {p}_i|}{{gt}_i} \right)  \,, \label{relative_error}
\end{equation}
where $gt_i$ and $p_i$ are the ground truth and prediction of the $i$-th case, and $N$ is the number of counting problems.

\textbf{Results.} As shown in Table~\ref{tab:count_accuracy}~\cite{xia2018dota}, RS-Agent outperforms GeoChat~\cite{kuckreja2023geochat} and LHRS-Bot~\cite{muhtar2024lhrs} on all metrics: 33.30\% absolute accuracy (vs.\ 17.65\% / 14.92\%), 75.98\% interval-matching (vs.\ 74.61\% / 60.78\%), and 0.28 relative error (vs.\ 0.65 / 0.56), yielding more precise and consistent predictions and confirming the robustness of RS-Agent for object counting. This advantage stems from the agent's ability to decompose counting into detection and aggregation steps, leveraging a specialized detector that end-to-end MLLMs lack.

\subsubsection{Visual Question Answering}
Visual Question Answering (VQA) requires understanding visual content and answering natural-language questions, jointly testing visual perception, language comprehension, and cross-modal reasoning. We adopt a multi-tool approach: RS-Agent invokes super-resolution, denoising, and GeoChat for different question types in RSVQA, retrieving strategies from $D_s$ via Task-Aware Retrieval.

\textbf{Datasets.}
We use RSVQA-LR~\cite{lobry2020rsvqa}, with 772 satellite images and 77{,}232 question–answer pairs across four question types: presence, comparison, rural/urban classification, and counting.

\textbf{Compared Methods.}
We compare RS-Agent with two groups of multimodal large language models: (1) general-purpose models trained on broad, non-specialized corpora, \eg, LLaVA-1.5~\cite{liu2023improved}, MiniGPTv2~\cite{chen2023minigpt}, InstructBLIP~\cite{dai2024instructblip}, mPLUG-Owl2~\cite{ye2023mplug} and QWen-VL-Chat~\cite{bai2023qwen} and (2) remote sensing-specific models fine-tuned on domain-specific remote sensing datasets, \eg, RSVQA~\cite{lobry2020rsvqa}, SkyEyeGPT~\cite{zhan2025skyeyegpt}, LHRS-Bot~\cite{muhtar2024lhrs} and GeoChat~\cite{kuckreja2023geochat}.

\textbf{Results.} As shown in Table~\ref{tab:performance_comparison}~\cite{liu2023improved,chen2023minigpt,dai2024instructblip,ye2023mplug}, with multi-tool invocation in a zero-shot setting, RS-Agent reaches 97.00\% on Rural/Urban, 91.07\% on Presence, and 90.58\% on Comparison, with an average of 90.88\%---the highest among all tested models, surpassing MLLMs such as GeoChat and LHRS-Bot on RSVQA-LR.
These results reflect the benefit of dynamically selecting and composing task-specific tools according to question type, a capability beyond end-to-end MLLMs.

\subsubsection{Scene Classification}
Scene classification categorizes satellite or aerial images into predefined scene types, testing the ability to capture spatial patterns, textures, and contextual cues. As the scene classifier, RS-Agent uses a Vision Transformer (ViT-B16)~\cite{dosovitskiy2020image} self-supervised pretrained on ImageNet~\cite{deng2009imagenet} and fine-tuned on RSSDIVCS~\cite{li2021learning} (70 categories, 800 images each at 256$\times$256, 80\% for training).

\textbf{Datasets.} We evaluate on RSSDIVCS~\cite{li2021learning}, AID~\cite{xia2017aid} (Google Earth aerial images, 30 classes), and UCMerced~\cite{yang2010bag} (2{,}100 images, 21 land use categories).

\textbf{Results.} As shown in Table~\ref{tab:scene_accuracy}~\cite{li2021learning,xia2017aid,yang2010bag}, RS-Agent achieves 98.00\% on RSSDIVCS~\cite{li2021learning} (vs.\ GeoChat 51.43\%, LHRS-Bot 63.85\%), 96.88\% on AID~\cite{xia2017aid} (vs.\ 72.03\% / 91.29\%), and 98.63\% on UCMerced~\cite{yang2010bag} (vs.\ 84.43\% / 96.63\%), consistently outperforming all baselines. The large margins confirm that correctly routing images to a fine-tuned domain classifier via the agent significantly outperforms the implicit classification ability of end-to-end MLLMs.

\begin{table*}[tb]
\centering
\captionsetup{justification=centering, labelsep=space, font=small, textfont=normal, labelfont=bf}
\caption{Ablation Study of Task-Aware Retrieval: Tool Execution Accuracy on Single-Tool and Multi-Tool Tasks.}
\label{tab:ablation_single_tool}
\resizebox{\textwidth}{!}{ 
\begin{tabular}{lccccccccc}
\toprule

Method  & \makecell{Object \\ Counting} & \makecell{Object \\ Detection} & \makecell{Land Use \\ Segmentation} & \makecell{Image \\ Captioning} & \makecell{Scene \\ Classification} & \makecell{Instance \\ Segmentation} & \makecell{Edge \\ Detection} & \makecell{Single-Tool \\ Accuracy} & \makecell{Multi-Tool \\ Accuracy}\\
\midrule
Baseline & 10.53\% & 15.79\% & 95.24\% & 85.71\% & 76.19\% & 100\% & 94.74\% & 69.06\% & 9.42\%\\
Baseline+$T_{\text{inf}}$ & 52.63\% & 36.84\% & 95.24\% & 95.24\% & 95.24\% & 100\% & 100\% & 82.01\% & 10.87\%\\
Baseline+$S_{\text{ret}}$ & 52.63\% & 31.58\% & 100\% & 80.95\% & 52.38\% & 94.74\% & 94.74\% & 73.38\% & 11.38\%\\
Baseline+$T_{\text{inf}}$+$S_{\text{ret}}$ & \textbf{100\%} & \textbf{84.21\%} & 90.48\% & \textbf{95.24\%} & \textbf{100\%} & 84.21\% & \textbf{100\%} & \textbf{93.53\%} & \textbf{90.62\%}\\

\bottomrule
\end{tabular}}
\end{table*}

\begin{figure*}[t!]
\begin{center}
\includegraphics[width=\linewidth]{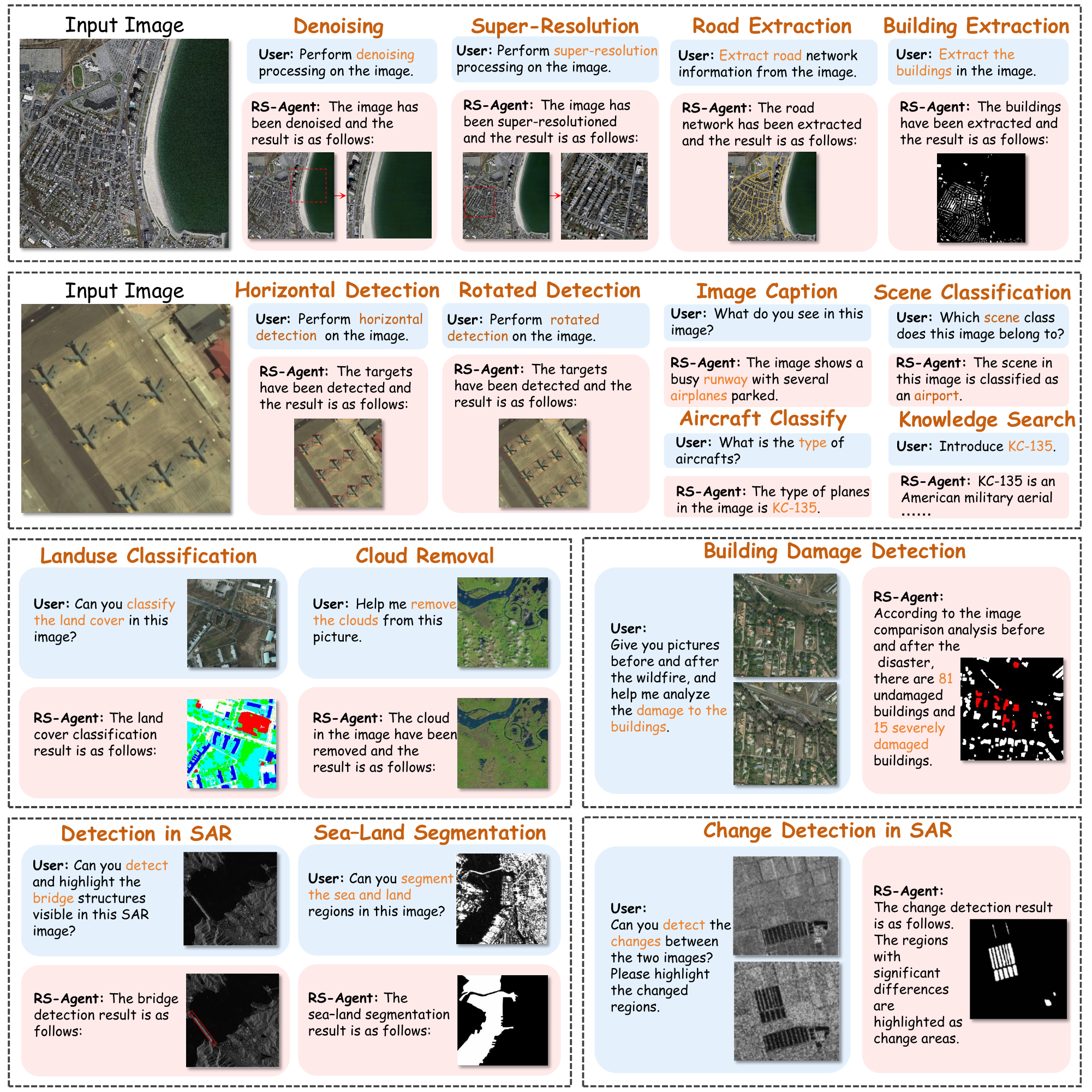}
\caption{Qualitative result of RS-Agent. This figure shows several key tools that highlight the core capabilities of RS-Agent. The input image represents the image input by the user. The blue box shows the user's request, and the red box shows the RS-Agent's reply.}
\label{fig:qualitative_result}
\end{center}
\vspace{-1.0em}
\end{figure*}

\subsection{Effectiveness of the Task-Aware Retrieval method}
Task-Aware Retrieval derives $g_s$ via Task Inference and Solution Retrieval (Section~\ref{sec:solution_space}). To assess its contribution, we ablate four progressively-built variants:
1) \textbf{Baseline}: RS-Agent without the Solution Space;
2) \textbf{Baseline+$T_{\text{inf}}$}: only Task Inference enabled;
3) \textbf{Baseline+$S_{\text{retr}}$}: only Solution Retrieval enabled, with $Q$ used directly as the retrieval query;
4) \textbf{Baseline+$T_{\text{inf}}$+$S_{\text{retr}}$}: the full RS-Agent.

\textbf{Evaluation tasks.} We test two task categories: single-tool tasks requiring one tool, and multi-tool tasks coupling several tools to assess complex workflows.

\textbf{Metric.} Task planning accuracy is reported as \textbf{Single-Tool Accuracy} (whether the correct tool is selected for single-tool tasks) and \textbf{Multi-Tool Accuracy} (whether all required tools are invoked in a valid order for multi-tool tasks). We additionally report end-to-end task success rates based on final-answer correctness in the Supplementary File~(Table~S1).

\textbf{Results.}
As shown in Table~\ref{tab:ablation_single_tool}, the Baseline achieves 69.06\% single-tool accuracy but only 9.42\% multi-tool task planning accuracy, as it frequently fails to invoke the correct tools (e.g., 10.53\% on Object Counting). Adding Task Inference or Solution Retrieval alone yields moderate single-tool gains (82.01\% and 73.38\%) with limited multi-tool improvement (10.87\% and 11.38\%). Combining both achieves the best performance: 93.53\% single-tool and 90.62\% multi-tool accuracy, confirming a strong synergistic effect. Notably, the underlying toolset is identical across all variants; the gains are entirely attributable to the orchestration mechanism. Multi-tool end-to-end task success rates are reported in Table~S1 (Supplementary File), and representative case studies are provided in Table~\ref{tab:case_study}.

To provide deeper insight into the roles of Task Inference and Solution Retrieval, we present representative case studies in Table~\ref{tab:case_study}. Each case compares the MLLM baseline (GeoChat~\cite{kuckreja2023geochat}) with the four RS-Agent ablation variants. These cases further illustrate that the core value of the agent lies in orchestration: even when powerful specialized tools are available, incorrect tool selection or improper execution ordering leads to task failure, underscoring the necessity of the orchestration mechanism.

\begin{table*}[!htbp]
\footnotesize
\centering
\caption{Case studies comparing the MLLM baseline (GeoChat) and RS-Agent ablation variants. Result: \gmark\ = correct, \rmark\ = incorrect.}
\label{tab:case_study}
\setlength{\tabcolsep}{3pt}
\renewcommand{\arraystretch}{1.15}
\begin{tabular}{p{3.2cm} l p{5.2cm} p{5.2cm} c}
\toprule
\textbf{Query} & \textbf{Variant} & \textbf{Tool Execution\textbackslash Output}
 & \textbf{Analysis} & \textbf{Result} \\
\midrule
\multirow[t]{10}{3.2cm}{\textit{``What targets are in this SAR image?''} \\ \mbox{(SAR Detection)} \\ \includegraphics[width=2.2cm]{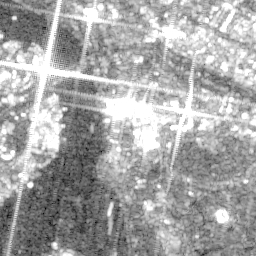}}
& MLLM
& ``The image shows some bright spots\ldots''
& No SAR-specific capability; vague description instead of detection results.
& \rmark \\[12pt]
& Baseline
& \texttt{optical\_det.}(img) $\to$ optical results
& Modality not distinguished; invoked optical tool on SAR data.
& \rmark \\[12pt]
& +$T_{\text{inf}}$
& \texttt{sar\_det.}(img) $\to$ correct detections
& Correctly invoked \texttt{sar\_det.}
& \gmark \\[9pt]
& +$S_{\text{retr}}$
& \texttt{sar\_det.}(img) $\to$ correct detections 
& Correctly invoked \texttt{sar\_det.}
& \gmark \\[9pt]
& +$T_{\text{inf}}$+$S_{\text{retr}}$
& \texttt{sar\_det.}(img) $\to$ correct detections
& Correctly invoked \texttt{sar\_det.}
& \gmark \\ [3pt]
\midrule
\multirow[t]{10}{3.2cm}{\textit{``Generate a landuse map from this image.''} \\ \mbox{(Land Use Seg.)} \\ \includegraphics[width=2.2cm]{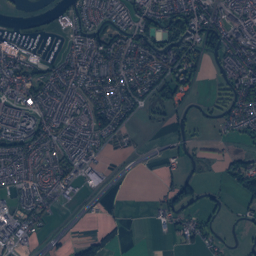}}
& MLLM
& ``urban areas, vegetation, roads\ldots''
& Cannot generate pixel-level segmentation; text-only output.
& \rmark \\[12pt]
& Baseline
& \texttt{landuse\_seg.}(img) $\to$ correct map
& Correctly invoked \texttt{landuse\_seg.}
& \gmark \\[9pt]
& +$T_{\text{inf}}$
& \texttt{sem\_seg.}(img) $\to$ wrong output
& Over-abstracted as ``Semantic Segmentation''; wrong tool.
& \rmark \\[12pt]
& +$S_{\text{retr}}$
& \texttt{landuse\_seg.}(img) $\to$ correct map
& Correctly invoked \texttt{landuse\_seg.}
& \gmark \\[9pt]
& +$T_{\text{inf}}$+$S_{\text{retr}}$
& \texttt{landuse\_seg.}(img) $\to$ correct map
& Correctly invoked \texttt{landuse\_seg.}
& \gmark \\[3pt]
\midrule
\multirow[t]{10}{3.2cm}{\textit{``Identify the plane type in this SAR image.''} \\ \mbox{(SAR Plane Classification)} \\ \includegraphics[width=2.2cm]{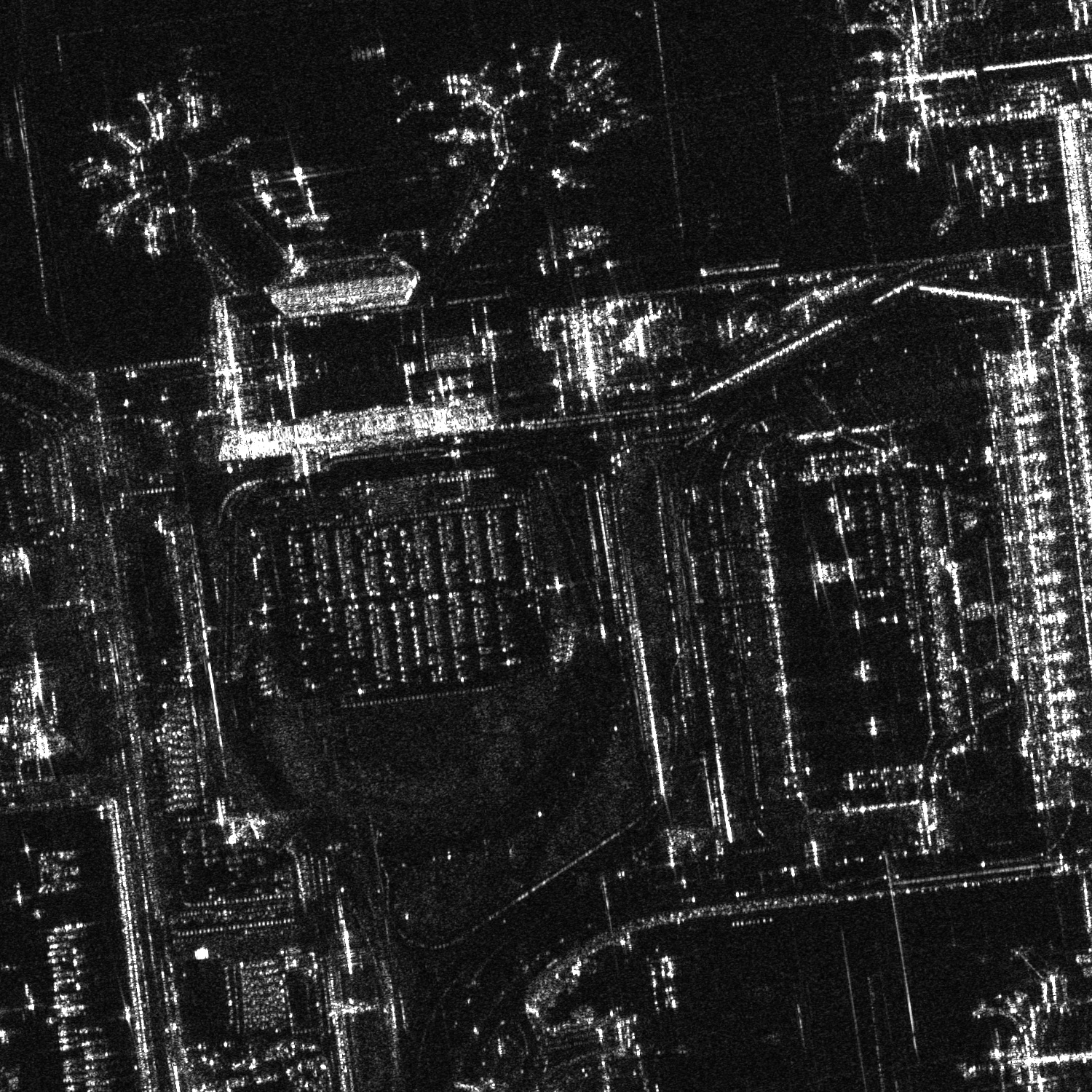}}
& MLLM
& ``I can see an aircraft\ldots''
& No SAR understanding capability; generic guess based on optical priors.
& \rmark \\[12pt]
& Baseline
& \texttt{optical\_plane\_type}(img) $\to$ incorrect
& Misidentified as optical task; applied optical classifier to SAR image.
& \rmark \\[12pt]
& +$T_{\text{inf}}$
& \texttt{sar\_plane\_type}(img) $\to$ ``A220.''
& Correctly invoked \texttt{sar\_plane\_type}
& \gmark \\[9pt]
& +$S_{\text{retr}}$
& \texttt{sar\_plane\_type}(img) $\to$ ``A220.''
& Correctly invoked \texttt{sar\_plane\_type}
& \gmark \\[9pt]
& +$T_{\text{inf}}$+$S_{\text{retr}}$
& \texttt{sar\_plane\_type}(img) $\to$ ``A220.''
& Correctly invoked \texttt{sar\_plane\_type}
& \gmark \\ [3pt]
\midrule
\multirow[t]{10}{3.2cm}{\textit{``Is there a circular residential building in the image?''} \\ \mbox{(RSVQA)} \\ \includegraphics[width=2.2cm]{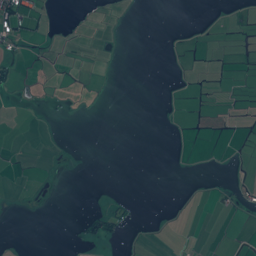}}
& MLLM
& ``Yes''
& Directly answered without preprocessing; missed fine-grained details in low-quality image.
& \rmark \\[12pt]
& Baseline
& \texttt{geochat}(Q, img) $\to$ ``Yes''
& Invoked \texttt{geochat} only; no image enhancement performed.
& \rmark \\[9pt]
& +$T_{\text{inf}}$
& \texttt{geochat}(Q, img) $\to$ ``Yes''
& Correctly inferred ``Presence''; but no preprocessing guidance retrieved.
& \rmark \\[9pt]
& +$S_{\text{retr}}$
& \texttt{geochat}(Q, img) $\to$ ``Yes''
& Retrieved wrong guidance.
& \rmark \\[9pt]
& +$T_{\text{inf}}$+$S_{\text{retr}}$
& \texttt{denoise}(img) $\to$ \texttt{super\_resolution}(img) $\to$ \texttt{geochat}(Q, img) $\to$ ``No''
& Correctly invoked multi-tool pipeline; image enhancement enabled accurate answer.
& \gmark \\ [3pt]
\bottomrule
\end{tabular}
\vspace{-1.5em}
\end{table*}

\begin{table*}[tb]
\centering
\captionsetup{justification=centering, labelsep=space, font=small, textfont=normal, labelfont=bf}
\caption{Win rates of LightRAG v.s. RS-Agent with DualRAG across two datasets and four evaluation dimensions.}
\label{tab:dualrag}
\resizebox{0.7\textwidth}{!}{%
\begin{tabular}{lcccccc}
\toprule
\multirow{2}{*}{Mode} & \multirow{2}{*}{Dimensions} & \multicolumn{2}{c}{\textbf{RSaircraft}} & \multicolumn{2}{c}{\textbf{Mix}} \\
\cmidrule(lr){3-4} \cmidrule(lr){5-6}
& & LightRAG & \textbf{RS-Agent(Ours)} & LightRAG & \textbf{RS-Agent(Ours)} \\
\midrule
\multirow{4}{*}{Local} & Comprehensiveness & 41.00\% & \textbf{58.00\%} & 48.40\% & \textbf{51.60\%} \\
 & Diversity & 43.40\% & \textbf{56.40\%} & 50.00\% & \textbf{50.00\%} \\
 & Empowerment & 43.40\% & \textbf{56.40\%} & \textbf{50.80\%} & 49.20\% \\
 & Overall & 42.80\%& \textbf{57.20\%} & \textbf{50.80\%} & 49.20\% \\
 
\midrule
\multirow{4}{*}{Global} & Comprehensiveness & 40.40\% & \textbf{59.60\%} & 48.80\% & \textbf{51.20\%} \\
 & Diversity & 43.60\% & \textbf{56.40\%} & 46.80\% & \textbf{53.20\%} \\
 & Empowerment & 39.20\% & \textbf{60.80\%} & \textbf{51.60\%} & 48.40\% \\
 & Overall & 39.60\% & \textbf{60.40\%} & 49.20\% & \textbf{50.80\%} \\

\midrule
\multirow{4}{*}{Hybrid} & Comprehensiveness & 36.80\% & \textbf{63.20\%} & 46.40\% & \textbf{53.60\%} \\
 & Diversity & 37.00\% & \textbf{62.00\%} & 44.80\% & \textbf{55.20\%} \\
 & Empowerment & 35.60\% & \textbf{64.40\%} & 47.60\% & \textbf{52.40\%} \\
 & Overall & 36.00\% & \textbf{64.00\%} & 47.20\% & \textbf{52.80\%} \\
\bottomrule
\end{tabular}}
\vspace{-1.0em}
\end{table*}

\textbf{Analysis.} These cases highlight three key findings. First, the end-to-end MLLM (GeoChat)~\cite{kuckreja2023geochat} consistently fails on tasks requiring precise quantitative outputs, modality-specific processing, or pixel-level predictions, as it can only produce text-based descriptions without invoking specialized tools. In contrast, RS-Agent leverages tool orchestration to deliver actionable, accurate results. Second, within RS-Agent's ablation variants: (1) Task Inference is most critical for queries with ambiguous task boundaries and cross-modal disambiguation (e.g., SAR vs.\ optical); (2) Solution Retrieval provides precise procedural guidance but may retrieve mismatched templates when used alone; (3) occasional over-abstraction by Task Inference (e.g., mapping ``landuse'' to ``segmentation'') is corrected when combined with Solution Retrieval. These observations confirm both the advantage of the agentic paradigm over end-to-end MLLMs and the synergistic design of Task-Aware Retrieval. Third, the RSVQA Presence~\cite{lobry2020rsvqa} case demonstrates RS-Agent's strength in multi-tool tasks: all single-component variants invoke only \texttt{geochat} and produce an incorrect answer, whereas the full system with both Task Inference and Solution Retrieval correctly plans a three-step pipeline (\texttt{denoise} $\to$ \texttt{super\_resolution} $\to$ \texttt{geochat}), improving image quality before visual reasoning and yielding the correct answer.

\vspace{-0.7em}
\subsection{Effectiveness of the DualRAG method}
\vspace{-0.7em}
To verify DualRAG's effectiveness, we compare RS-Agent with DualRAG against LightRAG~\cite{guo2024lightrag}.

\textbf{Datasets.} In this study, we utilize two datasets: the RSaircraft dataset, which we have specifically constructed for remote sensing applications, and the Mix dataset from the UltraDomain benchmark~\cite{qian2024memorag}, a general-domain RAG dataset. The RSaircraft dataset focuses on aircraft-related information, tailored for remote sensing tasks, and was built by systematically collecting and processing structured data from publicly available sources.

\textbf{Metric.}
Following LightRAG~\cite{guo2024lightrag}, GPT-4o-mini serves as an automated pairwise judge that selects the superior response between two candidates along four dimensions---Comprehensiveness, Diversity, Empowerment, and Overall quality---under three keyword usage modes: Local, Global, and Hybrid.

\textbf{Results.}
Table~\ref{tab:dualrag} reports win rates of LightRAG~\cite{guo2024lightrag} vs.\ RS-Agent with DualRAG. RS-Agent consistently surpasses LightRAG, with notably larger gains on RSaircraft, leading in all dimensions and modes—Local mode improves Comprehensiveness by 17.00\% (58.00\% vs.\ 41.00\%) and Hybrid mode shows the largest gaps (+26.40\% Comprehensiveness, +25.00\% Empowerment, +28.00\% Overall). Gains on Mix are smaller but consistent, confirming DualRAG's effectiveness and robustness across datasets and retrieval settings.

\vspace{-0.7em}
\subsection{Qualitative Results}
\vspace{-0.7em}
Figure~\ref{fig:qualitative_result} illustrates representative tasks showcasing RS-Agent's core capabilities. Each tool demonstrates strong performance, producing fluent, accurate, and contextually appropriate responses. Outputs are not only visually and semantically aligned with the input queries, but also exhibit robustness across various image types and modalities. Beyond accuracy, RS-Agent also excels in interpreting natural language instructions, even when they are underspecified or ambiguous. For example, a vague query such as “What do you see in this image?” is correctly interpreted as an image description task, and the system autonomously invokes the appropriate \texttt{image captioning} model to produce a detailed and relevant response.

This reflects strong task planning capabilities that bridge user intent and technical execution through semantic understanding rather than keyword matching. The modular toolkit allows seamless integration of new tools, ensuring long-term scalability and adaptability across diverse remote sensing scenarios.

\vspace{-0.7em}
\section{Conclusion}
\vspace{-0.7em}

We propose RS-Agent, an AI agent tailored for remote sensing that seamlessly integrates diverse tools to support a wide range of tasks across multiple imaging modalities. The proposed architecture enables user query understanding, complex task planning, tool execution, and context-aware memory. Our Task-Aware Retrieval method significantly improves task planning accuracy, enhancing both efficiency and reliability in complex scenarios. In addition, the DualRAG mechanism provides access to specialized domain knowledge, enabling RS-Agent to handle complex technical queries. Extensive experiments demonstrate that RS-Agent consistently outperforms state-of-the-art multimodal large language models across multiple remote sensing benchmarks. Moreover, RS-Agent supports efficient integration of new tools while retaining previously learned capabilities. Overall, RS-Agent shows strong potential to streamline remote sensing workflows, reduce reliance on domain experts, and enable more accessible, high-quality analysis in real-world applications.

Despite its strong performance, RS-Agent has several limitations. The Task Inference module can occasionally over-abstract user queries, leading to incorrect tool selection, and the agent sometimes generates malformed tool arguments. Reinforcement learning from execution feedback or supervised fine-tuning on tool-calling traces could address these issues. In multi-tool workflows, upstream errors can propagate to downstream tools, calling for intermediate quality checks and adaptive re-planning. On the evaluation side, the current metrics focus on tool-selection correctness and final-answer accuracy but do not diagnose intermediate step quality at a fine-grained level; developing more comprehensive workflow-level evaluation schemes is an important direction for future work. Additionally, the Knowledge and Solution Databases currently have limited coverage and rely on manual curation; automated web-based knowledge collection could broaden their scope. Finally, domain-specific fine-tuning or distillation could reduce the system's dependence on large proprietary LLMs.


\Acknowledgements{This work was supported by the National Natural Science Foundation of China (Grant No.62301063 and No.42501551), and ``the funds for National Key Laboratory of Inertial Measurement".}

\bibliographystyle{unsrt}
\bibliography{ref}

@article{wang2024survey,
  title={A survey on large language model based autonomous agents},
  author={Wang, Lei and Ma, Chen and Feng, Xueyang and Zhang, Zeyu and Yang, Hao and Zhang, Jingsen and Chen, Zhiyuan and Tang, Jiakai and Chen, Xu and Lin, Yankai and others},
  journal={Frontiers of Computer Science},
  volume={18},
  number={6},
  pages={1--26},
  year={2024},
  publisher={Springer}
}

@article{qin2023tool,
  title={Tool learning with foundation models},
  author={Qin, Yujia and Hu, Shengding and Lin, Yankai and Chen, Weize and Ding, Ning and Cui, Ganqu and Zeng, Zheni and Huang, Yufei and Xiao, Chaojun and Han, Chi and others},
  journal={arXiv preprint arXiv:2304.08354},
  year={2023}
}

@article{shapira2023clever,
  title={Clever hans or neural theory of mind? stress testing social reasoning in large language models},
  author={Shapira, Natalie and Levy, Mosh and Alavi, Seyed Hossein and Zhou, Xuhui and Choi, Yejin and Goldberg, Yoav and Sap, Maarten and Shwartz, Vered},
  journal={arXiv preprint arXiv:2305.14763},
  year={2023}
}

@article{kenneweg2024retrieval,
  title={Retrieval Augmented Generation Systems: Automatic Dataset Creation, Evaluation and Boolean Agent Setup},
  author={Kenneweg, Tristan and Kenneweg, Philip and Hammer, Barbara},
  journal={arXiv preprint arXiv:2403.00820},
  year={2024}
}

@article{gao2023retrieval,
  title={Retrieval-augmented generation for large language models: A survey},
  author={Gao, Yunfan and Xiong, Yun and Gao, Xinyu and Jia, Kangxiang and Pan, Jinliu and Bi, Yuxi and Dai, Yi and Sun, Jiawei and Wang, Haofen},
  journal={arXiv preprint arXiv:2312.10997},
  year={2023}
}

@article{kuckreja2023geochat,
  title={Geochat: Grounded large vision-language model for remote sensing},
  author={Kuckreja, Kartik and Danish, Muhammad Sohail and Naseer, Muzammal and Das, Abhijit and Khan, Salman and Khan, Fahad Shahbaz},
  journal={arXiv preprint arXiv:2311.15826},
  year={2023}
}

@article{dosovitskiy2020image,
  title={An image is worth 16x16 words: Transformers for image recognition at scale},
  author={Dosovitskiy, Alexey and Beyer, Lucas and Kolesnikov, Alexander and Weissenborn, Dirk and Zhai, Xiaohua and Unterthiner, Thomas and Dehghani, Mostafa and Minderer, Matthias and Heigold, Georg and Gelly, Sylvain and others},
  journal={arXiv preprint arXiv:2010.11929},
  year={2020}
}

@article{du2023tree,
  title={Tree-GPT: Modular Large Language Model Expert System for Forest Remote Sensing Image Understanding and Interactive Analysis},
  author={Du, Siqi and Tang, Shengjun and Wang, Weixi and Li, Xiaoming and Guo, Renzhong},
  journal={arXiv preprint arXiv:2310.04698},
  year={2023}
}

@article{chen2023minigpt,
  title={Minigpt-v2: large language model as a unified interface for vision-language multi-task learning},
  author={Chen, Jun and Zhu, Deyao and Shen, Xiaoqian and Li, Xiang and Liu, Zechun and Zhang, Pengchuan and Krishnamoorthi, Raghuraman and Chandra, Vikas and Xiong, Yunyang and Elhoseiny, Mohamed},
  journal={arXiv preprint arXiv:2310.09478},
  year={2023}
}

@article{cui2025overview,
  title={Overview of AI and communication for 6G network: Fundamentals, challenges, and future research opportunities},
  author={Cui, Qimei and You, Xiaohu and Wei, Ni and Nan, Guoshun and Zhang, Xuefei and Zhang, Jianhua and Lyu, Xinchen and Ai, Ming and Tao, Xiaofeng and Feng, Zhiyong and others},
  journal={Science China Information Sciences},
  volume={68},
  number={7},
  pages={171301},
  year={2025},
  publisher={Springer}
}

@article{shabbir2025thinkgeo,
  title={Thinkgeo: Evaluating tool-augmented agents for remote sensing tasks},
  author={Shabbir, Akashah and Munir, Muhammad Akhtar and Dudhane, Akshay and Sheikh, Muhammad Umer and Khan, Muhammad Haris and Fraccaro, Paolo and Moreno, Juan Bernabe and Khan, Fahad Shahbaz and Khan, Salman},
  journal={arXiv preprint arXiv:2505.23752},
  year={2025}
}

@article{wang2020word,
  title={Word--sentence framework for remote sensing image captioning},
  author={Wang, Qi and Huang, Wei and Zhang, Xueting and Li, Xuelong},
  journal={IEEE Transactions on Geoscience and Remote Sensing},
  volume={59},
  number={12},
  pages={10532--10543},
  year={2020},
  publisher={IEEE}
}

@article{chen2025towards,
  title={Towards wireless native big AI model: the mission and approach differ from large language model},
  author={Chen, Zirui and Zhang, Zhaoyang and Liu, Chenyu and Xing, Ziqing},
  journal={Science China Information Sciences},
  volume={68},
  number={7},
  pages={170303},
  year={2025},
  publisher={Springer}
}

@article{lobry2020rsvqa,
  title={RSVQA: Visual question answering for remote sensing data},
  author={Lobry, Sylvain and Marcos, Diego and Murray, Jesse and Tuia, Devis},
  journal={IEEE Transactions on Geoscience and Remote Sensing},
  volume={58},
  number={12},
  pages={8555--8566},
  year={2020},
  publisher={IEEE}
}

@article{hu2025rsgpt,
  title={Rsgpt: A remote sensing vision language model and benchmark},
  author={Hu, Yuan and Yuan, Jianlong and Wen, Congcong and Lu, Xiaonan and Liu, Yu and Li, Xiang},
  journal={ISPRS Journal of Photogrammetry and Remote Sensing},
  volume={224},
  pages={272--286},
  year={2025},
  publisher={Elsevier}
}

@article{zhan2025skyeyegpt,
  title={Skyeyegpt: Unifying remote sensing vision-language tasks via instruction tuning with large language model},
  author={Zhan, Yang and Xiong, Zhitong and Yuan, Yuan},
  journal={ISPRS Journal of Photogrammetry and Remote Sensing},
  volume={221},
  pages={64--77},
  year={2025},
  publisher={Elsevier}
}

@article{touvron2023llama,
  title={Llama: Open and efficient foundation language models},
  author={Touvron, Hugo and Lavril, Thibaut and Izacard, Gautier and Martinet, Xavier and Lachaux, Marie-Anne and Lacroix, Timoth{\'e}e and Rozi{\`e}re, Baptiste and Goyal, Naman and Hambro, Eric and Azhar, Faisal and others},
  journal={arXiv preprint arXiv:2302.13971},
  year={2023}
}

@misc{langchain,
  author = {Chase, Harrison},
  title = {{LangChain}: Building applications with {LLM}s through composability},
  year = {2022},
  howpublished = {\url{https://github.com/langchain-ai/langchain}},
  note = {Version 0.1}
}

@misc{yolov8_github,
  author = {{Ultralytics}},
  title = {{YOLOv8}: State-of-the-art real-time object detection},
  year = {2023},
  howpublished = {\url{https://github.com/ultralytics/ultralytics}},
  note = {Version 8.0}
}

@article{li2021learning,
  title={Learning deep cross-modal embedding networks for zero-shot remote sensing image scene classification},
  author={Li, Yansheng and Zhu, Zhihui and Yu, Jin-Gang and Zhang, Yongjun},
  journal={IEEE Transactions on Geoscience and Remote Sensing},
  volume={59},
  number={12},
  pages={10590--10603},
  year={2021},
  publisher={IEEE}
}

@article{xia2017aid,
  title={AID: A benchmark data set for performance evaluation of aerial scene classification},
  author={Xia, Gui-Song and Hu, Jingwen and Hu, Fan and Shi, Baoguang and Bai, Xiang and Zhong, Yanfei and Zhang, Liangpei and Lu, Xiaoqiang},
  journal={IEEE Transactions on Geoscience and Remote Sensing},
  volume={55},
  number={7},
  pages={3965--3981},
  year={2017},
  publisher={IEEE}
}

@inproceedings{yang2010bag,
  title={Bag-of-visual-words and spatial extensions for land-use classification},
  author={Yang, Yi and Newsam, Shawn},
  booktitle={Proceedings of the 18th SIGSPATIAL international conference on advances in geographic information systems},
  pages={270--279},
  year={2010}
}

@article{muhtar2024lhrs,
  title={LHRS-Bot: Empowering Remote Sensing with VGI-Enhanced Large Multimodal Language Model},
  author={Muhtar, Dilxat and Li, Zhenshi and Gu, Feng and Zhang, Xueliang and Xiao, Pengfeng},
  journal={arXiv preprint arXiv:2402.02544},
  year={2024}
}

@inproceedings{xia2018dota,
  title={DOTA: A large-scale dataset for object detection in aerial images},
  author={Xia, Gui-Song and Bai, Xiang and Ding, Jian and Zhu, Zhen and Belongie, Serge and Luo, Jiebo and Datcu, Mihai and Pelillo, Marcello and Zhang, Liangpei},
  booktitle={Proceedings of the IEEE conference on computer vision and pattern recognition},
  pages={3974--3983},
  year={2018}
}

@article{liu2023improved,
  title={Improved baselines with visual instruction tuning},
  author={Liu, Haotian and Li, Chunyuan and Li, Yuheng and Lee, Yong Jae},
  journal={arXiv preprint arXiv:2310.03744},
  year={2023}
}

@article{dai2024instructblip,
  title={Instructblip: Towards general-purpose vision-language models with instruction tuning},
  author={Dai, Wenliang and Li, Junnan and Li, Dongxu and Tiong, Anthony Meng Huat and Zhao, Junqi and Wang, Weisheng and Li, Boyang and Fung, Pascale N and Hoi, Steven},
  journal={Advances in Neural Information Processing Systems},
  volume={36},
  year={2024}
}

@article{ye2023mplug,
  title={mplug-owl: Modularization empowers large language models with multimodality},
  author={Ye, Qinghao and Xu, Haiyang and Xu, Guohai and Ye, Jiabo and Yan, Ming and Zhou, Yiyang and Wang, Junyang and Hu, Anwen and Shi, Pengcheng and Shi, Yaya and others},
  journal={arXiv preprint arXiv:2304.14178},
  year={2023}
}

@article{bai2023qwen,
  title={Qwen-vl: A frontier large vision-language model with versatile abilities},
  author={Bai, Jinze and Bai, Shuai and Yang, Shusheng and Wang, Shijie and Tan, Sinan and Wang, Peng and Lin, Junyang and Zhou, Chang and Zhou, Jingren},
  journal={arXiv preprint arXiv:2308.12966},
  year={2023}
}

@article{tao2023general,
  author={Tao, Chao and Xiao, Rong and Wang, Yuze and Qi, Ji and Li, Haifeng},
  journal={IEEE Journal of Selected Topics in Applied Earth Observations and Remote Sensing}, 
  title={A General Transitive Transfer Learning Framework for Cross-Optical Sensor Remote Sensing Image Scene Understanding}, 
  year={2023},
  volume={16},
  number={},
  pages={4248-4260},
  doi={10.1109/JSTARS.2023.3269852}}

@article{bashmal2023visual,
  title={Visual question generation from remote sensing images},
  author={Bashmal, Laila and Bazi, Yakoub and Melgani, Farid and Ricci, Riccardo and Al Rahhal, Mohamad M and Zuair, Mansour},
  journal={IEEE Journal of Selected Topics in Applied Earth Observations and Remote Sensing},
  volume={16},
  pages={3279--3293},
  year={2023},
  publisher={IEEE}
}

@article{ruan2023tptu,
  title={Tptu: Task planning and tool usage of large language model-based ai agents},
  author={Ruan, Jingqing and Chen, Yihong and Zhang, Bin and Xu, Zhiwei and Bao, Tianpeng and Du, Guoqing and Shi, Shiwei and Mao, Hangyu and Zeng, Xingyu and Zhao, Rui},
  journal={arXiv preprint arXiv:2308.03427},
  year={2023}
}

@inproceedings{deng2009imagenet,
  title={Imagenet: A large-scale hierarchical image database},
  author={Deng, Jia and Dong, Wei and Socher, Richard and Li, Li-Jia and Li, Kai and Fei-Fei, Li},
  booktitle={2009 IEEE conference on computer vision and pattern recognition},
  pages={248--255},
  year={2009},
  organization={Ieee}
}

@article{chen2024bge,
  title={Bge m3-embedding: Multi-lingual, multi-functionality, multi-granularity text embeddings through self-knowledge distillation},
  author={Chen, Jianlv and Xiao, Shitao and Zhang, Peitian and Luo, Kun and Lian, Defu and Liu, Zheng},
  journal={arXiv preprint arXiv:2402.03216},
  year={2024}
}

@article{jegou2022faiss,
  title={Faiss: Similarity search and clustering of dense vectors library},
  author={J{\'e}gou, Herv{\'e} and Douze, Matthijs and Johnson, Jeff and Hosseini, Lucas and Deng, Chengqi},
  journal={Astrophysics Source Code Library},
  pages={ascl--2210},
  year={2022}
}

@article{edge2024local,
  title={From local to global: A graph rag approach to query-focused summarization},
  author={Edge, Darren and Trinh, Ha and Cheng, Newman and Bradley, Joshua and Chao, Alex and Mody, Apurva and Truitt, Steven and Metropolitansky, Dasha and Ness, Robert Osazuwa and Larson, Jonathan},
  journal={arXiv preprint arXiv:2404.16130},
  year={2024}
}

@article{guo2024lightrag,
  title={Lightrag: Simple and fast retrieval-augmented generation},
  author={Guo, Zirui and Xia, Lianghao and Yu, Yanhua and Ao, Tian and Huang, Chao},
  journal={arXiv preprint arXiv:2410.05779},
  volume={2},
  number={3},
  year={2024}
}

@inproceedings{guo2024remote,
  title={Remote sensing chatgpt: Solving remote sensing tasks with chatgpt and visual models},
  author={Guo, Haonan and Su, Xin and Wu, Chen and Du, Bo and Zhang, Liangpei and Li, Deren},
  booktitle={IGARSS 2024-2024 IEEE International Geoscience and Remote Sensing Symposium},
  pages={11474--11478},
  year={2024},
  organization={IEEE}
}

@article{li2024unirs,
  title={UniRS: Unifying Multi-temporal Remote Sensing Tasks through Vision Language Models},
  author={Li, Yujie and Xu, Wenjia and Li, Guangzuo and Yu, Zijian and Wei, Zhiwei and Wang, Jiuniu and Peng, Mugen},
  journal={arXiv preprint arXiv:2412.20742},
  year={2024}
}

@article{luo2024skysensegpt,
  title={Skysensegpt: A fine-grained instruction tuning dataset and model for remote sensing vision-language understanding},
  author={Luo, Junwei and Pang, Zhen and Zhang, Yongjun and Wang, Tingzhu and Wang, Linlin and Dang, Bo and Lao, Jiangwei and Wang, Jian and Chen, Jingdong and Tan, Yihua and others},
  journal={arXiv preprint arXiv:2406.10100},
  year={2024}
}

@article{liu2024change,
  title={Change-agent: Towards interactive comprehensive remote sensing change interpretation and analysis},
  author={Liu, Chenyang and Chen, Keyan and Zhang, Haotian and Qi, Zipeng and Zou, Zhengxia and Shi, Zhenwei},
  journal={IEEE Transactions on Geoscience and Remote Sensing},
  year={2024},
  publisher={IEEE}
}

@article{shi2017can,
  title={Can a machine generate humanlike language descriptions for a remote sensing image?},
  author={Shi, Zhenwei and Zou, Zhengxia},
  journal={IEEE Transactions on Geoscience and Remote Sensing},
  volume={55},
  number={6},
  pages={3623--3634},
  year={2017},
  publisher={IEEE}
}

@inproceedings{radford2021learning,
  title={Learning transferable visual models from natural language supervision},
  author={Radford, Alec and Kim, Jong Wook and Hallacy, Chris and Ramesh, Aditya and Goh, Gabriel and Agarwal, Sandhini and Sastry, Girish and Askell, Amanda and Mishkin, Pamela and Clark, Jack and others},
  booktitle={International conference on machine learning},
  pages={8748--8763},
  year={2021},
  organization={PmLR}
}

@article{sun2023eva,
  title={Eva-clip: Improved training techniques for clip at scale},
  author={Sun, Quan and Fang, Yuxin and Wu, Ledell and Wang, Xinlong and Cao, Yue},
  journal={arXiv preprint arXiv:2303.15389},
  year={2023}
}

@article{sun2021pbnet,
  title={PBNet: Part-based convolutional neural network for complex composite object detection in remote sensing imagery},
  author={Sun, Xian and Wang, Peijin and Wang, Cheng and Liu, Yingfei and Fu, Kun},
  journal={ISPRS Journal of Photogrammetry and Remote Sensing},
  volume={173},
  pages={50--65},
  year={2021},
  publisher={Elsevier}
}

@article{li2021deep,
  title={A deep translation (GAN) based change detection network for optical and SAR remote sensing images},
  author={Li, Xinghua and Du, Zhengshun and Huang, Yanyuan and Tan, Zhenyu},
  journal={ISPRS Journal of Photogrammetry and Remote Sensing},
  volume={179},
  pages={14--34},
  year={2021},
  publisher={Elsevier}
}

@article{wang2022unetformer,
  title={UNetFormer: A UNet-like transformer for efficient semantic segmentation of remote sensing urban scene imagery},
  author={Wang, Libo and Li, Rui and Zhang, Ce and Fang, Shenghui and Duan, Chenxi and Meng, Xiaoliang and Atkinson, Peter M},
  journal={ISPRS Journal of Photogrammetry and Remote Sensing},
  volume={190},
  pages={196--214},
  year={2022},
  publisher={Elsevier}
}

@article{zhu2022land,
  title={Land-use/land-cover change detection based on a Siamese global learning framework for high spatial resolution remote sensing imagery},
  author={Zhu, Qiqi and Guo, Xi and Deng, Weihuan and Shi, Sunan and Guan, Qingfeng and Zhong, Yanfei and Zhang, Liangpei and Li, Deren},
  journal={ISPRS Journal of Photogrammetry and Remote Sensing},
  volume={184},
  pages={63--78},
  year={2022},
  publisher={Elsevier}
}

@article{reed2022generalist,
  title={A generalist agent},
  author={Reed, Scott and Zolna, Konrad and Parisotto, Emilio and Colmenarejo, Sergio Gomez and Novikov, Alexander and Barth-Maron, Gabriel and Gimenez, Mai and Sulsky, Yury and Kay, Jackie and Springenberg, Jost Tobias and others},
  journal={arXiv preprint arXiv:2205.06175},
  year={2022}
}

@inproceedings{park2023generative,
  title={Generative agents: Interactive simulacra of human behavior},
  author={Park, Joon Sung and O'Brien, Joseph and Cai, Carrie Jun and Morris, Meredith Ringel and Liang, Percy and Bernstein, Michael S},
  booktitle={Proceedings of the 36th annual acm symposium on user interface software and technology},
  pages={1--22},
  year={2023}
}

@article{brown2020language,
  title={Language models are few-shot learners},
  author={Brown, Tom and Mann, Benjamin and Ryder, Nick and Subbiah, Melanie and Kaplan, Jared D and Dhariwal, Prafulla and Neelakantan, Arvind and Shyam, Pranav and Sastry, Girish and Askell, Amanda and others},
  journal={Advances in neural information processing systems},
  volume={33},
  pages={1877--1901},
  year={2020}
}

@inproceedings{zhu2024rs,
  title={RS-AGENT: Large Language Models Guided Agent System for Remote Sensing Image Generation},
  author={Zhu, Lingxuan and Wu, Jiaji and Wang, Biao and Zhang, Guoshuai and Wang, Jiarong and Chen, Shaohong and Tan, Mingzhou},
  booktitle={IGARSS 2024-2024 IEEE International Geoscience and Remote Sensing Symposium},
  pages={7020--7024},
  year={2024},
  organization={IEEE}
}

@article{wang2024earthvqanet,
  title={EarthVQANet: Multi-task visual question answering for remote sensing image understanding},
  author={Wang, Junjue and Ma, Ailong and Chen, Zihang and Zheng, Zhuo and Wan, Yuting and Zhang, Liangpei and Zhong, Yanfei},
  journal={ISPRS Journal of Photogrammetry and Remote Sensing},
  volume={212},
  pages={422--439},
  year={2024},
  publisher={Elsevier}
}

@article{dong2024changeclip,
  title={ChangeCLIP: Remote sensing change detection with multimodal vision-language representation learning},
  author={Dong, Sijun and Wang, Libo and Du, Bo and Meng, Xiaoliang},
  journal={ISPRS Journal of Photogrammetry and Remote Sensing},
  volume={208},
  pages={53--69},
  year={2024},
  publisher={Elsevier}
}

@inproceedings{gao2023precise,
  title={Precise zero-shot dense retrieval without relevance labels},
  author={Gao, Luyu and Ma, Xueguang and Lin, Jimmy and Callan, Jamie},
  booktitle={Proceedings of the 61st Annual Meeting of the Association for Computational Linguistics (Volume 1: Long Papers)},
  pages={1762--1777},
  year={2023}
}

@article{chan2024rq,
  title={Rq-rag: Learning to refine queries for retrieval augmented generation},
  author={Chan, Chi-Min and Xu, Chunpu and Yuan, Ruibin and Luo, Hongyin and Xue, Wei and Guo, Yike and Fu, Jie},
  journal={arXiv preprint arXiv:2404.00610},
  year={2024}
}

@article{tang2024multihop,
  title={Multihop-rag: Benchmarking retrieval-augmented generation for multi-hop queries},
  author={Tang, Yixuan and Yang, Yi},
  journal={arXiv preprint arXiv:2401.15391},
  year={2024}
}

@article{zhu2025knowledge,
  title={Knowledge Graph-Guided Retrieval Augmented Generation},
  author={Zhu, Xiangrong and Xie, Yuexiang and Liu, Yi and Li, Yaliang and Hu, Wei},
  journal={arXiv preprint arXiv:2502.06864},
  year={2025}
}

@article{guo2025deepseek,
  title={Deepseek-r1: Incentivizing reasoning capability in llms via reinforcement learning},
  author={Guo, Daya and Yang, Dejian and Zhang, Haowei and Song, Junxiao and Zhang, Ruoyu and Xu, Runxin and Zhu, Qihao and Ma, Shirong and Wang, Peiyi and Bi, Xiao and others},
  journal={arXiv preprint arXiv:2501.12948},
  year={2025}
}

@article{yang2024qwen2,
  title={Qwen2. 5 technical report},
  author={Yang, An and Yang, Baosong and Zhang, Beichen and Hui, Binyuan and Zheng, Bo and Yu, Bowen and Li, Chengyuan and Liu, Dayiheng and Huang, Fei and Wei, Haoran and others},
  journal={arXiv preprint arXiv:2412.15115},
  year={2024}
}

@article{qian2024memorag,
  title={Memorag: Moving towards next-gen rag via memory-inspired knowledge discovery},
  author={Qian, Hongjin and Zhang, Peitian and Liu, Zheng and Mao, Kelong and Dou, Zhicheng},
  journal={arXiv preprint arXiv:2409.05591},
  volume={1},
  year={2024}
}

@article{xi2025rise,
  title={The rise and potential of large language model based agents: A survey},
  author={Xi, Zhiheng and Chen, Wenxiang and Guo, Xin and He, Wei and Ding, Yiwen and Hong, Boyang and Zhang, Ming and Wang, Junzhe and Jin, Senjie and Zhou, Enyu and others},
  journal={Science China Information Sciences},
  volume={68},
  number={2},
  pages={121101},
  year={2025},
  publisher={Springer}
}

@article{wu2023autogen,
  title={Autogen: Enabling next-gen llm applications via multi-agent conversation},
  author={Wu, Qingyun and Bansal, Gagan and Zhang, Jieyu and Wu, Yiran and Li, Beibin and Zhu, Erkang and Jiang, Li and Zhang, Xiaoyun and Zhang, Shaokun and Liu, Jiale and others},
  journal={arXiv preprint arXiv:2308.08155},
  year={2023}
}

@article{feng2025earthagent,
  title={Earth-agent: Unlocking the full landscape of earth observation with agents},
  author={Feng, Peilin and Lv, Zhutao and Ye, Junyan and Wang, Xiaolei and Huo, Xinjie and Yu, Jinhua and Xu, Wanghan and Zhang, Wenlong and Bai, Lei and He, Conghui and others},
  journal={arXiv preprint arXiv:2509.23141},
  year={2025}
}

@article{wang2024robust, title={Robust fast adaptation from adversarially explicit task distribution generation}, author={Wang, Cheems and Lv, Yiqin and Mao, Yixiu and Qu, Yun and Xu, Yi and Ji, Xiangyang}, journal={arXiv preprint arXiv:2407.19523}, year={2024} }

@article{lv2024theoretical, title={Theoretical investigations and practical enhancements on tail task risk minimization in meta learning}, author={Lv, Yiqin and Wang, Qi and Liang, Dong and Xie, Zheng}, journal={arXiv preprint arXiv:2410.22788}, year={2024} }

\clearpage

\setcounter{section}{0}
\renewcommand{\thesection}{\Alph{section}}
\renewcommand{\thesubsection}{\thesection.\arabic{subsection}}
\renewcommand{\thesubsubsection}{\thesubsection.\arabic{subsubsection}}
\setcounter{figure}{0}
\setcounter{table}{0}
\renewcommand{\thefigure}{S\arabic{figure}}
\renewcommand{\thetable}{S\arabic{table}}

\begin{center}
{\large\bfseries Supplementary File}
\end{center}

\section{Detailed Prompt Designs}
\label{appendix:prompts}

This section provides the complete prompt templates used in RS-Agent's Central Controller for task inference and keyword extraction. These prompts are carefully designed to incorporate remote sensing domain knowledge and guide the LLM to make accurate task planning decisions.

\subsection{Task Inference Prompt}
The Task Inference prompt guides the Central Controller to analyze user queries and infer the appropriate task types from the supported task list. The detailed prompt is illustrated as follows.

\begin{lstlisting}[style=customprompt, breaklines=true, breakindent=0pt, columns=fullflexible]
(*@\textbf{Role Definition}@*)
You are RS-Agent, a remote sensing expert responsible for inferring the required task types from a user query. Your goal is to analyze the user's remote-sensing analysis intent and select the minimal and sufficient set of task types from the supported task list.

(*@\textbf{Remote Sensing Reasoning Cues}@*)
When interpreting the query, consider the following remote-sensing factors:
1.Sensor/modality (optical, SAR, thermal, etc., or unknown);
2.Spatial granularity (object-level/scene-level/region-level);
3.Temporal setting (single-date, pre-post, or multi-temporal);
4.Analytical operation (detection, segmentation/mapping, classification, change analysis, severity/impact assessment, counting/statistics, or description/QA).

(*@\textbf{Knowledge-Assisted Inference (Optional)}@*)
If the task type cannot be confidently inferred from the query alone, you may first perform a knowledge search to retrieve relevant remote-sensing background information, and then re-evaluate the task type based on the acquired knowledge.
Prefer high-level analytical tasks over low-level processing steps, and avoid redundant or implied tasks.

(*@\textbf{Consistency Check}@*)
Before producing the final output, ensure that:
1.All task types are selected from supported_tasks;
2.The task set is minimal, non-redundant, and sufficient to satisfy the user's analytical intent.

(*@\textbf{Inputs:}@*)
- supported_tasks: {supported_tasks}
- user_query: {Q}

(*@\textbf{Output (JSON only):}@*)
{{"task_type":["task1","task2","task3"]}}
\end{lstlisting}

\subsection{Keyword Extraction Prompt}
The Keyword Extraction prompt is used in the DualRAG mechanism to extract important keywords from user queries with weighted importance scores. This enables more effective knowledge retrieval by prioritizing remote sensing domain-specific terminology. The format of the keyword generation prompt is shown below:

\begin{lstlisting}[style=customprompt, breaklines=true, breakindent=0pt, columns=fullflexible]
(*@\textbf{Role Definition}@*)
You are a remote sensing keyword extractor for RAG. From the query, extract concise, searchable keywords and assign importance (1-10). Prioritize remote-sensing professional terminology and abbreviations; omit trivial words.

(*@\textbf{Extraction Principles}@*)
When extracting keywords, follow these principles:
- Prioritize remote-sensing professional terminology, standard abbreviations, and well-defined technical concepts;
- Omit trivial, conversational, or low-information words;

(*@\textbf{Remote Sensing Prioritization Order}@*)
Prefer keywords in the following descending order of importance: Task or analysis target; Analytical or processing concept; Sensor, modality, or platform; Indices, variables, or domain-specific attributes; Temporal setting ; Spatial scale or resolution; Area of interest or location.

(*@\textbf{Importance Scoring Rules}@*)
Assign an importance score to each keyword:
9-10: core constraints essential for correct retrieval;
6-8: important qualifiers that refine the query;
3-5: broad contextual information;
Omit keywords with negligible retrieval value.

(*@\textbf{Inputs:}@*)
- user_query: {Q}
(*@\textbf{Output (JSON only):}@*)
{"keywords":[{"term":"...","importance":1-10}]}
\end{lstlisting}

\section{Multi-Tool End-to-End Task Success Rate}
\label{appendix:multi-tool-e2e}

To complement the task planning accuracy reported in Table~6 of the main text, we further evaluate multi-tool end-to-end task success rate, measured by final-answer correctness on the RSVQA-LR benchmark~\cite{lobry2020rsvqa}. As shown in Table~\ref{tab:multi_tool_e2e}, the Baseline, +$T_{\text{inf}}$, and +$S_{\text{retr}}$ variants achieve comparable end-to-end accuracy (84.24\%, 84.34\%, and 84.57\%, respectively), with notably low Rural/Urban accuracy (20\%--29\%), indicating that partial orchestration components alone are insufficient for complex multi-tool workflows. In contrast, the full RS-Agent (+$T_{\text{inf}}$+$S_{\text{retr}}$) achieves 90.88\% overall, with Rural/Urban accuracy jumping to 97\%, demonstrating that the synergistic combination of Task Inference and Solution Retrieval is essential for planning effective multi-step tool pipelines.

\begin{table}[!htbp]
\centering
\captionsetup{justification=centering, labelsep=space, font=small, textfont=normal, labelfont=bf}
\caption{Multi-tool end-to-end task success rate on RSVQA-LR.}
\label{tab:multi_tool_e2e}
\begin{tabular}{lcccc}
\toprule
Method & Rural/Urban & Presence & Compare & Avg. \\
\midrule
Baseline & 20.00\% & 87.23\% & 83.62\% & 84.24\% \\
+$T_{\text{inf}}$ & 23.00\% & 87.58\% & 83.48\% & 84.34\% \\
+$S_{\text{retr}}$ & 29.00\% & 87.58\% & 83.73\% & 84.57\% \\
+$T_{\text{inf}}$+$S_{\text{retr}}$ & \textbf{97.00\%} & \textbf{91.07\%} & \textbf{90.58\%} & \textbf{90.88\%} \\
\bottomrule
\end{tabular}
\end{table}

\section{Complete Task Planning Accuracy}
\label{appendix:full-llm-ablation}

Table~2 in the main text reports task planning accuracy on 10 representative tasks. Table~\ref{tab:ablation_llm_full} below provides the complete results for all 18 evaluation tasks. The ranking across LLMs remains consistent with the main-text subset.

\begin{table*}[!htbp]
\centering
\captionsetup{justification=centering, labelsep=space, font=small, textfont=normal, labelfont=bf}
\resizebox{\textwidth}{!}{
\begin{threeparttable}
\caption{Complete Task Planning Accuracy of RS-Agent with Different LLMs (18 evaluation tasks).}
\label{tab:ablation_llm_full}
\begin{tabular}{lccccccccc}
\toprule
\multirow{2}{*}{Task} & \multicolumn{3}{c}{ChatGPT} & \multicolumn{2}{c}{LLaMa 3.1} & \multicolumn{3}{c}{Qwen2.5} & \multicolumn{1}{c}{DeepSeek} \\
\cmidrule(lr){2-4} \cmidrule(lr){5-6} \cmidrule(lr){7-9} \cmidrule(lr){10-10}
 & \makecell{3.5-turbo-1106 \\ (87.71t/s)} & \makecell{3.5-turbo \\ (65.03t/s)} & \makecell{4o-mini \\ (58.87t/s)} & \makecell{8B \\ (100.78t/s)} & \makecell{70B \\ (17.71t/s)} & \makecell{14B \\ (69.61t/s)} & \makecell{32B \\ (36.77t/s)} & \makecell{72B \\ (16.24t/s)} & \makecell{r1:70B \\ (18.25t/s)} \\
\midrule
Cloud Removal          & 95.00\% & 95.00\% & 100\% & 100\% & 100\% & 100\% & 95.00\% & 100\% & 100\% \\
Image Dehazing         & 30.00\% & 95.00\% & 100\% & 100\% & 100\% & 100\% & 100\% & 100\% & 75.00\% \\
Super Resolution       & 100\% & 100\% & 100\% & 0\% & 100\% & 100\% & 100\% & 100\% & 95.00\% \\
Denoising              & 90.00\% & 100\% & 100\% & 100\% & 100\% & 100\% & 100\% & 100\% & 90.00\% \\
Image Captioning       & 55.00\% & 45.00\% & 90.00\% & 15.00\% & 60.00\% & 70.00\% & 80.00\% & 80.00\% & 10.00\% \\
Object Detection       & 75.00\% & 60.00\% & 95.00\% & 30.00\% & 90.00\% & 90.00\% & 85.00\% & 100\% & 85.00\% \\
Optical Plane Classif. & 100\% & 100\% & 100\% & 100\% & 100\% & 100\% & 100\% & 100\% & 95.00\% \\
Scene Classification   & 20.00\% & 90.00\% & 100\% & 80.00\% & 90.00\% & 90.00\% & 100\% & 100\% & 50.00\% \\
SAR Detection          & 30.00\% & 100\% & 100\% & 75.00\% & 95.00\% & 100\% & 100\% & 100\% & 100\% \\
SAR Plane Classif.     & 100\% & 100\% & 100\% & 100\% & 100\% & 100\% & 100\% & 100\% & 90.00\% \\
Knowledge Search       & 100\% & 100\% & 100\% & 100\% & 80.00\% & 100\% & 100\% & 100\% & 10.00\% \\
Building Damage Det.   & 100\% & 100\% & 100\% & 100\% & 100\% & 95.00\% & 100\% & 100\% & 100\% \\
Building Extraction    & 10.00\% & 70.00\% & 100\% & 55.00\% & 100\% & 100\% & 100\% & 100\% & 100\% \\
Road Extraction        & 15.00\% & 55.00\% & 100\% & 65.00\% & 100\% & 100\% & 100\% & 100\% & 100\% \\
Horizontal Detection   & 20.00\% & 55.00\% & 100\% & 95.00\% & 100\% & 100\% & 100\% & 100\% & 100\% \\
Rotated Detection      & 15.00\% & 35.00\% & 100\% & 85.00\% & 90.00\% & 100\% & 100\% & 100\% & 100\% \\
Semantic Segmentation  & 60.00\% & 100\% & 100\% & 80.00\% & 100\% & 100\% & 100\% & 100\% & 80.00\% \\
Land Use Classif.      & 15.00\% & 100\% & 100\% & 75.00\% & 100\% & 100\% & 100\% & 95.00\% & 95.00\% \\
\midrule
Average Accuracy       & 57.22\% & 82.50\% & 99.17\% & 75.28\% & 94.72\% & 96.94\% & 97.78\% & 98.61\% & 81.94\% \\
\bottomrule
\end{tabular}
\begin{tablenotes}
\footnotesize
\item[1]``B'' denotes the number of parameters in billions.
\item[2] Numbers in parentheses (t/s) indicate model inference speed in tokens per second on NVIDIA RTX4090 GPUs.
\end{tablenotes}
\end{threeparttable}}
\end{table*}

\section{Tool List}
\label{appendix:tool-list}
Table~\ref{tab:tool_list} provides a complete overview of the 27 tools integrated into the RS-Agent framework, organized by modality.

\begin{table*}[!htbp]
\centering
\captionsetup{justification=centering, labelsep=space, font=small, textfont=normal, labelfont=bf}
\caption{Overview of the RS-Agent Toolkit. Tools are organized by sensing modality and analysis level.}
\label{tab:tool_list}
\footnotesize
\begin{tabular}{lllp{7.5cm}}
\toprule
\textbf{Modality} & \textbf{Level} & \textbf{Tool Name} & \textbf{Tool Description} \\
\midrule
\multirow{18}{*}{\makecell[l]{Optical}}
  & \multirow{5}{*}{\makecell[l]{Low-Level}}
  & Super Resolution ($\times$2) & Enhance spatial resolution of optical imagery (2$\times$ upscaling) \\
  && Super Resolution ($\times$4) & Enhance spatial resolution of optical imagery (4$\times$ upscaling) \\
  && Image Denoising & Remove noise from optical images \\
  && Cloud Removal & Remove cloud cover from optical imagery \\
  && Dehazing & Remove haze effects from optical imagery \\
\cmidrule(lr){2-4}
  & \multirow{13}{*}{\makecell[l]{High-Level}}
  & Visual Question Answering & Answer natural-language questions about optical images \\
  && Optical Object Detection & Detect objects in optical imagery \\
  && Object Cropping & Crop and extract regions of interest from images \\
  && Aircraft Type Recognition & Classify aircraft types in optical images \\
  && Scene Classification & Classify scene categories of optical images \\
  && Aircraft Counting & Count the number of aircraft in optical images \\
  && Building Damage Assessment & Assess building damage levels from optical images \\
  && Building Extraction & Extract building footprints from optical imagery \\
  && Road Extraction & Extract road networks from optical imagery \\
  && Land-Cover Classification & Classify land cover types in optical images \\
  && Knowledge-Based Reasoning & Retrieve and reason over remote sensing knowledge \\
  && Horizontal Object Detection & Detect objects with horizontal bounding boxes \\
  && Rotated Object Detection & Detect objects with oriented bounding boxes \\
\midrule
\multirow{9}{*}{\makecell[l]{SAR}}
  & \multirow{1}{*}{\makecell[l]{Low-Level}}
  & Despeckling & Remove speckle noise from SAR images \\
\cmidrule(lr){2-4}
  & \multirow{7}{*}{\makecell[l]{High-Level}}
  & SAR Object Detection & Detect targets in SAR imagery \\
  && Aircraft Type Recognition & Classify aircraft types in SAR images \\
  && Rotated Detection & Detect objects with oriented bounding boxes in SAR imagery \\
  && Land-Cover Classification & Classify land cover types in SAR imagery \\
  && Semantic Segmentation & Perform pixel-level semantic segmentation on SAR images \\
  && Sea-Land Segmentation & Segment sea and land regions in SAR images \\
  && Change Detection & Detect changes between multi-temporal SAR images \\
  && Target Recognition & Recognize and classify targets in SAR images \\
\bottomrule
\end{tabular}
\end{table*}
\section{Multi-Tool Collaboration Example}
\label{appendix:multi-tool}

\begin{figure}[!htbp]
\begin{center}
\includegraphics[width=0.8\linewidth]{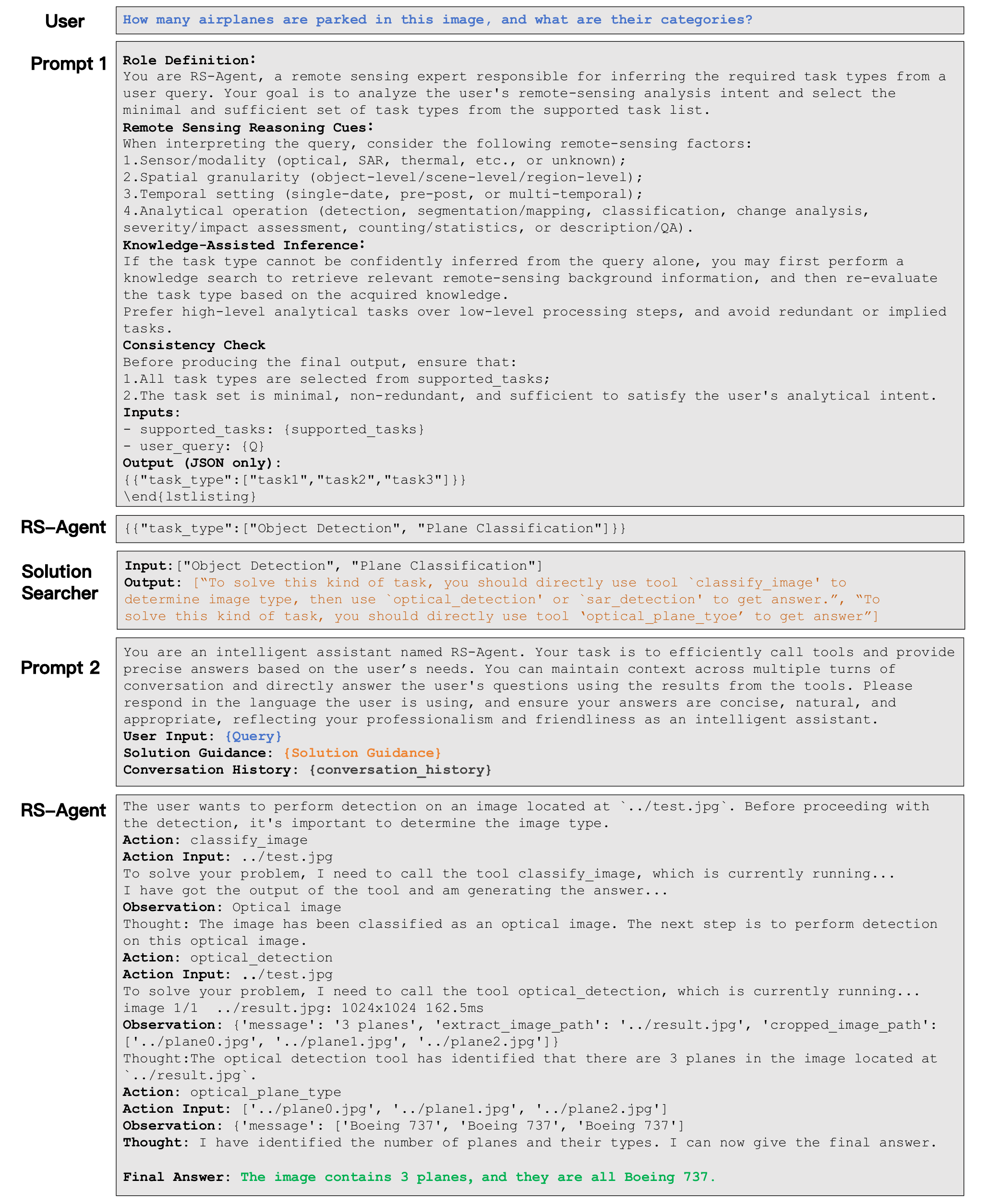}
\caption{Illustration of the RS-Agent workflow. Prompt1 infers the task type, the Solution Searcher retrieves task-specific guidance, and Prompt2 plans tool execution to produce the final answer.}
\label{fig:multi_tool}
\end{center}
\end{figure}

This section demonstrates RS-Agent's capability to handle complex queries that require coordinated invocation of multiple tools. Figure~\ref{fig:multi_tool} illustrates a complete workflow where the agent first infers the task type through Prompt1, retrieves task-specific guidance from the Solution Searcher, and then plans the tool execution sequence through Prompt2 to produce the final answer.

\section{User Interface}
\label{appendix:interface}

\begin{figure}[b!]
\begin{center}
\includegraphics[width=0.7\linewidth]{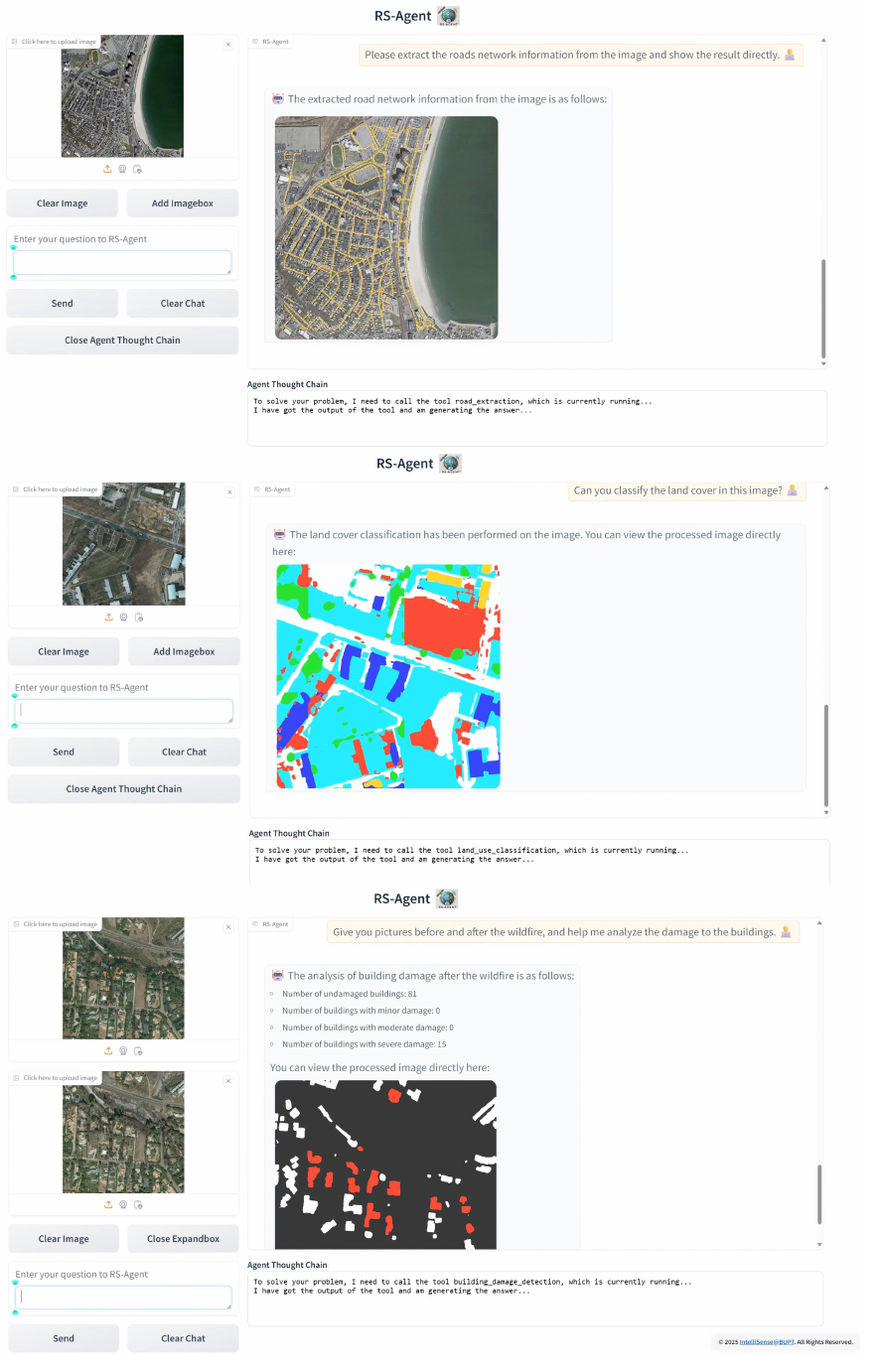}
\caption{The user interface of the RS-Agent. Users can upload remote sensing images and interpret remote sensing data through natural language. The example shows object detection and building damage detection task.}
\label{fig:interface}
\end{center}
\end{figure}

This section presents the user interface of RS-Agent, which provides an intuitive platform for users to interact with the system through natural language. As shown in Figure~\ref{fig:interface}, users can upload remote sensing images and perform various tasks such as object detection and building damage assessment by simply describing their analysis needs in plain language, without requiring expertise in remote sensing or programming.

\end{document}